\documentclass{article}
\usepackage{kr}

\usepackage{times}
\usepackage{soul}
\usepackage{url}
\usepackage[hidelinks]{hyperref}
\usepackage[utf8]{inputenc}
\usepackage[small]{caption}
\usepackage{graphicx}
\usepackage{amsmath}
\usepackage{amsthm}
\usepackage{booktabs}
\usepackage{algorithm}
\usepackage{algorithmic}
\usepackage{mathtools}

\usepackage{booktabs}
\usepackage{multirow}

\usepackage{subcaption}

\usepackage{enumitem}

\usepackage{ amssymb }

\usepackage{todonotes}

\usepackage{mymacros}

\newtheorem{example}{Example}
\newtheorem{theorem}{Theorem}
\newtheorem{lemma}{Lemma}
\newtheorem{definition}{Definition}
\newtheorem{transformation}{Transformation}

\title{Grounding Rule-Based Argumentation Using Datalog\thanks{This is the authors’ self-archived copy, including proofs, of a paper accepted for publication at the KR-25 conference.
}}

\author{%
Martin Diller$^1$\and
Sarah Alice Gaggl$^1$\and
Philipp Hanisch$^2$\and
Giuseppina	Monterosso$^3$ \and 
Fritz	Rauschenbach$^1$\\
\affiliations
$^1$Logic Programming and Argumentation Group, TU Dresden, Germany \\
$^2$Knowledge-Based Systems Group, TU Dresden, Germany\\
$^3$DIMES - University of Calabria, Italy\\
\emails
\{martin.diller, sarah.gaggl, philipp.hanisch1\}@tu-dresden.de, giusy.monterosso@dimes.unical.it, fritz.rauschenbach@proton.me
}

\begin{document}

\maketitle

\begin{abstract}

ASPIC+ is one of the main general frameworks for rule-based argumentation for AI. 
Although first-order rules are commonly used in ASPIC+ examples, most existing approaches to reason over rule-based 
argumentation only support propositional rules. 
To enable reasoning over first-order instances, a preliminary grounding step
 is required. 
  As groundings can lead to an exponential increase in the size of the input theories, intelligent procedures are needed. However, there is a lack
  of dedicated solutions for ASPIC+.
  Therefore, we propose an intelligent grounding procedure 
  that keeps the size of the grounding manageable while preserving the correctness of the reasoning process. 
  To this end, we translate the first-order ASPIC+ instance into a Datalog program and query a Datalog engine to obtain ground substitutions 
  to perform the grounding of rules and contraries. Additionally, we propose simplifications specific to the ASPIC+ formalism
  to avoid grounding of rules that have no influence on the reasoning process. Finally, we performed an empirical
  evaluation of a prototypical implementation to show scalability.

\end{abstract}

\section{Introduction} 

Rule-based argumentation formalisms such as ASPIC+~\cite{Prakken10} and the closely related Assumption-based Argumentation 
(ABA)~\cite{BondarenkoTK93,BondarenkoDKT97} provide a means to model and reason about complex scenarios involving conflicting information 
(see also~\cite{BesnardGHMPST14} for an overview). 
While ASPIC+ is a general abstract framework, a particularly relevant instance is what we refer to as (concrete) rule-based ASPIC+, 
also known as the logic-programming or Horn variant of ASPIC+. 
As the name suggests, in this variant, conflicts within knowledge bases composed of strict and defeasible Horn rules are resolved by inspecting arguments and counterarguments constructed using these rules. 
In particular, since rule-based ASPIC+ captures Answer Set Programming (ASP)~\cite{MarekT99,Niemela99} under the stable semantics, it also offers an argument-based characterization of ASP~\cite{ModgilP18,BondarenkoDKT97}. 
On the other hand, rule-based ASPIC+ supports a more expressive syntax than (normal) ASP—such as a generalized contrary relation and the inclusion of defeasible rules—and it relies on argumentation semantics~\cite{Dung95} rather than distinguished Herbrand models.

Although ASPIC+ and ASP share overlapping conceptual roots, ASPIC+ lags behind ASP in terms of the modelling constructs available in practical systems (see e.g.~\cite{GebserS16}). In particular, 
a key advantage of ASP systems is their support for first-order variables in Horn rules, enabling concise representations through universal quantification—a crucial feature for knowledge representation and reasoning. 
It is similarly natural for ASPIC+ instances to make use of first-order logic-programming-like rules, which support more compact and general modeling~\cite{ModgilP18}. 
Nevertheless, a concrete syntax for first-order ASPIC+ is often left unspecified. More critically, most existing computational techniques for reasoning in rule-based ASPIC+ assume that all rules are ground instances, i.e., propositional Horn rules~\cite{Aspforaspic1,OdekerkenLBWJ23,LehtonenOWJ24}. 

Relatedly, recent years have seen growing interest in developing sophisticated solvers for argumentation-exemplified by the International Competition on Computational Models of Argument (ICCMA), which has been held biennially since 2015 and in 2023 included, for the first time, a structured argumentation track focused on (flat) ABA frameworks~\cite{jarvisaloln25}, which can be captured in ASPIC+~\cite{ModgilP18}. Nevertheless, there remains a lack of benchmarks for rule-based instances derived from real-world problems. This gap may be partly due to the absence of effective methods for converting first-order rule knolwedge bases into their propositional counterparts.

For all these reasons, support for grounding—i.e., replacing variables with constants—is essential to broaden the applicability of rule-based ASPIC+. However, as is well known from related fields, grounding can be a major computational bottleneck: naive grounding strategies may cause exponential blow-up by generating numerous irrelevant propositional rules (see e.g.~\cite{BesinHW23}). 
In ASP, this problem has been extensively addressed, resulting in efficient grounders such as \gringo~\cite{gringo} and \pname{i-DLV}~\cite{calimeri2017dlv}. 
In contrast, no comparable solutions exist specifically for rule-based argumentation.

In this work, we address this gap by proposing a grounding approach for ASPIC+ that leverages engines for Datalog—a declarative language widely used for database querying and reasoning, which can also be seen as the fragment of ASP that excludes negation-as-failure (see e.g.~\cite{EastT00}). 
Specifically, our contributions are:

\noindent

\begin{enumerate}
\item We define a first-order syntax for rule-based argumentation within the ASPIC+ framework.
\item We propose an intelligent grounding procedure for first-order rule-based ASPIC+ that minimizes the grounding size while preserving the outcomes of the reasoning process. Our approach builds upon query answering in the context of Datalog, albeit incorporating optimizations specific to ASPIC+.
\item We present a prototype grounder, \angry, which utilizes the Datalog engine \nemo~\cite{ivliev2024nemo}.
    \item We demonstrate the feasibility of our approach through empirical evaluation in three distinct scenarios. In particular, we evaluate \angry~as part of a ground+solve pipeline for ASPIC+, using the system \aspforaspic~\cite{Aspforaspic1}, and compare it to \argtp~\cite{CalegariOPS22}, the only existing system supporting a first-order ASPIC+-like syntax. To assess its efficiency in the context of ASP, we also compare \angry—used as an ASP grounder—against the state-of-the-art ASP grounder \gringo.
    \end{enumerate}

\noindent

\textbf{Related Work.} The main system we are aware of that supports a first-order ASPIC+-like syntax is \pname{Arg2P}~\cite{CalegariOPS22}, implemented in TUProlog~\cite{CiattoCO21}. However, it has been shown to be impractical for large instances~\cite{legalBench}. Another notable logic programming-inspired argumentation formalism that supports first-order variables is DeLP~\cite{GarciaS04}. However, DeLP employs a different semantics~\cite{GarciaPS20}, and to our knowledge, no recent systematic empirical evaluations of DeLP systems exist. Regarding ASPIC+, and similarly for ABA, existing computational work primarily focuses on propositional instances (e.g. also~\cite{CravenT16,DillerGG21,LehtonenWJ21,PopescuW23,LehtonenR0W23,LehtonenRT0W24} for ABA).

Existing translations between ABA and ASPIC+~\cite{Heyninck19}, as well as between ASP and ABA~\cite{abaLpEquivalence}, suggest an alternative approach to grounding ASPIC+ frameworks. This would involve:  
(1) translating an ASPIC+ framework into an ASP program,  
(2) applying an ASP grounder, and  
(3) translating the resulting grounded ASP program back into an ASPIC+ theory.  However, this translation-based method faces several limitations.  
Regarding step (1): existing translations assume propositional ASPIC+.   
Regarding step (2): ASP grounders are tailored for the stable model semantics, whereas ASPIC+ reasoning may require alternative semantics. For instance, the translations in~\cite{abaLpEquivalence} utilize not only stable but also 3-valued stable, well-founded, regular, and ideal semantics. These ASP grounders (e.g., \gringo) often include hard-coded optimizations for stable models, which are not easily configurable or extensible.  
Regarding step (3): ASP grounders are typically designed with the aim of ultimately producing sets of ground atoms (answer sets, via an ASP solver), while ASPIC+ often requires sets of ground arguments (extensions). Optimizations targeted at the former may hinder the latter—as we show in our study.  Perhaps most importantly, even if translation-based grounding approaches for ASPIC+ could be developed, our method offers a simpler, more transparent, and modular alternative. 

The main goal of this work is to lay the theoretical foundation for grounding ASPIC+, and to demonstrate its practicality through our prototype grounder, \angry. The system is built on top of the Datalog engine \nemo~\cite{ivliev2024nemo}, though in principle, any of the several systems supporting Datalog on offer could be used (see e.g. also~\cite{NenovPMHWB15,JordanSS16,UrbaniJK16}), including also ASP grounders like \pname{Gringo} and \pname{i-DLV}. This underlines our motivation for choosing Datalog as a foundation: it offers a simpler and more focused computational model than full ASP, and our method does not require the additional features ASP provides.

At the same time, several of the techniques we propose for optimizing ASPIC+ grounding via Datalog are inspired by established ASP grounding techniques. As previously mentioned, since rule-based ASPIC+ subsumes ASP under stable semantics, our grounding procedure also offers a viable alternative for grounding ASP programs via translation to ASPIC+. However, this remains a secondary benefit. Our primary objective is to enable effective grounding for rule-based ASPIC+.  Finally, we note that the ASP literature also explores alternative grounding mechanisms as those that interleave grounding and solving (e.g.~\cite{WeinzierlTF20}), which fall outside the scope of this work.

\section{Background}\label{sec:02}

\subsection{Propositional rule-based argumentation in the ASPIC+ framework}\label{subsec-prop-aspic}

As we indicated in the introduction, among the various instantiations of ASPIC+, the most widely used is arguably that capturing logic 
programming-style argumentation. We call it \emph{concrete} rule-based ASPIC+ or simply rule-based ASPIC+ for short 
(see also~\cite{DungT14}).  Based on~\cite{ModgilP18} we start by defining a convenient syntax for the propositional variant of this instance. 
We assume as given a countably infinite set of propositional atoms, denoted by $\lan$.  

\begin{definition}  An \emph{\argpr} is a tuple $\asppv = \aspp$. $\con$ is a finite set of expressions $\crul{\stmv}{\bodc}$ mapping
  an $\stmv \in \lan$ to its set of \emph{contraries} $\{\bodc\} \subseteq \lan$.  $\ruls$ is a finite set of \emph{strict} rules, having 
  the form $\srul{\bodv}{\hedv}$ with $\bodv \cup \{\hedv\} \subseteq \lan$. $\ruld$ is a finite set of \emph{defeasible} rules, which have 
  the form $\drul{\namv}{\bodv}{\hedv}$ with $\bodv \cup \{\namv,\hedv\} \subseteq \lan$.  
  Finally, $\kbsa \subseteq \lan$ and
   $\kbso \subseteq \lan$, for which $\kbsa \cap \kbso = \emptyset$, are the set of \emph{facts} and \emph{assumptions} 
   respectively\footnote{These are called axiom and ordinary premisses in the ASPIC+ framework; we use the notions ``facts" 
   and ``assumptions" to denote the concrete case in which they are propositional atoms.}. 
   Both $\kbsa$ and $\kbso$ are also finite. The set of rules of $\asppv$ is then $\rul = \ruls \cup \ruld$ and the knowledge base $\kbs = \kbsa \cup \kbso$.        
\end{definition}

By minor abuse of notation, we will treat $\con$ both as a set of expressions as well as the obvious (partial) function 
$\lan \mapsto \ps{\lan}$ it induces. For simplicity we will also often represent rules $\rulv = \drul{\namv}{\bodv}{\hedv} \in \ruld$ 
as $\rulv' = \drult{\bodv}{\hedv} \in \ruld$, with the name $\namv$ of a defeasible rule (often implicitly) associated to $\rulv'$ via a 
naming function $\pfu: \ruld \mapsto \lan$, \ie~$\pfua{\rulv'} = \namv$.    
Then, in particular, we can use $\rulv = \arul{\bodv}{\hedv}$ to denote an arbitrary rule $\rulv \in \rul$.  We will often also omit 
the curly braces for sets and rather list their elements, e.g. $\rulv = \arul{\bodr}{\hedv}$ for the rule above with $\bodv = \{\bodr\}$.
Arguments in ASPIC+ involve deriving claims from facts and assumptions via the strict and defeasible rules:

\begin{definition}\label{def:args}  The set of \emph{arguments} of an \argpr~$\asppv = \aspp$ is defined inductively as 
follows: {\bf i }) if $\stmv \in \kbsa \cup \kbso$, then $\argv = \stmv$ is an argument with \emph{conclusion} 
$\conc{\argv} = \stmv$, \emph{premisses} $\pre{\argv} = \stmv$, and \emph{rules} $\rulsa{\argv} = \emptyset$; {\bf ii}) 
if $\argv_1,\ldots,\argv_m$ are arguments and $\rulv = \arul{\conc{\argv_1},\ldots,\conc{\argv_m}}{\hedv} \in \rul$, 
then $\argv = \aarg{\argv_1,\ldots,\argv_m}{\hedv}$ is an argument with $\conc{\argv} = \hedv$, 
$\pre{\argv} = \bigcup_{1 \leq i \leq m}\pre{\argv_i}$, $\rulsa{\argv} = \bigcup_{1 \leq i \leq m}\rulsa{\argv_i} \cup \{\rulv\}$, 
and \emph{top-rule} $\trl{\argv} = \rulv$.  There are no other arguments than those defined by i) and ii).  \end{definition}

\noindent Note that by definition arguments are always finite, \ie~they are constructed by finite application of rules.  
In this work we will often also extend notation to sets in the obvious manner and without explicit definition; then, e.g. 
for a set of arguments $\argsv$, $\conc{\argsv} = \bigcup_{\argv \in \argsv} \conc{\argv}$.  Conflicts between arguments are captured by the notion of attack:

\begin{definition}\label{def:asp-attack} Let $\asppv = \aspp$ be an \argpr. An argument $\argv$ of $\asppv$ \emph{attacks} an argument $\argv'$ of $\asppv$ iff $\argv$ undercuts, rebuts, or undermines $\argv'$ where {\bf i)} $\argv$ \emph{undercuts} $\argv'$ (on $\rulv$) iff $\conc{\argv} \in \cona{\pfua{\rulv}}$ for a $\rulv \in \rulsa{\argv'} \cap \ruld$, {\bf ii)} $\argv$ \emph{rebuts} $\argv'$ (on $\drult{\bodv}{\hedv}$) iff $\conc{\argv} \in \cona{\hedv}$ for a $\drult{\bodv}{\hedv} \in \rulsa{\argv'} \cap \ruld$, and {\bf iii)} $\argv$ \emph{undermines} $\argv'$ (on $\stmv$) iff $\conc{\argv}  \in \cona{\stmv}$ for a $\stmv \in \pre{\argv} \cap \kbso$.
\end{definition}

\noindent Note that attacks are always on some \emph{defeasible element} from $\defe{\argv'} = (\pre{\argv'} \cap \kbso) \cup (\rulsa{\argv'} \cap \ruld)$.  
On the other hand, if we define $\defe{\rulv} = \{\namv, \hedv\}$ and $\cona{\rulv} = \cona{\defe{\rulv}} = \cona{\namv} \cup \cona{\hedv}$ for a $\rulv = \drul{\namv}{\bodv}{\hedv} \in \ruld$, while $\cona{\rulv} = \cona{\defe{\rulv'}} = \defe{\rulv'}  = \emptyset$ for $\rulv' \in \ruls$, then attacks are always through some element in $\cona{\defe{\argv'}} = \bigcup_{\defv \in \defe{\argv'}} \cona{\defv}$.  

To evaluate ASPIC+ theories, these are traditionally translated into argumentation graphs~\cite{Dung95}: 

\begin{definition} 
An (abstract) \emph{argumentation framework} (or \emph{graph}) (AF) is tuple $(\aargs,\aatr)$ where $\aargs$ is a
 set of (abstract) arguments and $\aatr \subseteq \aargs \times \aargs$ the attack relation.
\end{definition}

\begin{definition} 
For an AF $(\aargs,\aatr)$, a set $\aargsv \subseteq \aargs$ is {\bf i)} \emph{conflict-free}  iff there are
 no $\aargv,\aargv' \in \aargsv$ \stt~$(\aargv,\aargv') \in \aatr$, {\bf ii)} \emph{admissible} (shorthand: $\adm$) iff $\aargsv$ is 
 conflict free and every $\aargv \in \aargsv$ is defended by $\aargsv$, where $\aargsv$ \emph{defends} $\aargv$ if for 
 every $(\aargv',\aargv) \in \aatr$, $\aargsv$ also \emph{attacks} $\aargv'$, \ie~there is a $\aargv'' \in \aargsv$ 
 \stt~$(\aargv'',\aargv') \in \aatr$, {\bf iii)} \emph{complete} ($\com$) iff $\aargsv$ is admissible and includes 
 every $\aargv \in \aargs$ it defends, {\bf iv)} \emph{grounded} ($\grd$) iff $\aargsv$ is subset-minimal among the  
 complete sets, {\bf v)} \emph{preferred} ($\prf$) iff $\aargsv$ is subset-maximal among the  complete 
 sets, {\bf vi)} \emph{stable} ($\stb$) iff $\aargsv$ is admissible and attacks every $\aargv \in \aargs \setminus \aargsv$.
\end{definition}

\begin{definition} 
The \emph{AF defined by the \argpr} $\asppv$ is the argumentation graph $(\arts{\asppv},\atts{\asppv})$ 
where $\arts{\asppv}$ are all the arguments of $\asppv$ and $\atts{\asppv}$ is the attack relation among arguments induced 
by $\asppv$.   $\stmsv \subseteq \lan$ is credulously (skeptically) \emph{acceptable} for $\semv \in \semlist$ iff there exists 
a $\semv$-extension\footnote{An \emph{extension} is a set of arguments that satisfies the criteria of the semantics.} (for 
all $\semv$-extensions) $\argsv \subseteq \arts{\asppv}$, $\stmsv \subseteq \conc{\argsv}$. 
\end{definition}

\begin{figure}[h]
    \centering
    \begin{subfigure}[b]{0.22\textwidth}
        \centering
        \includegraphics[width=\linewidth]{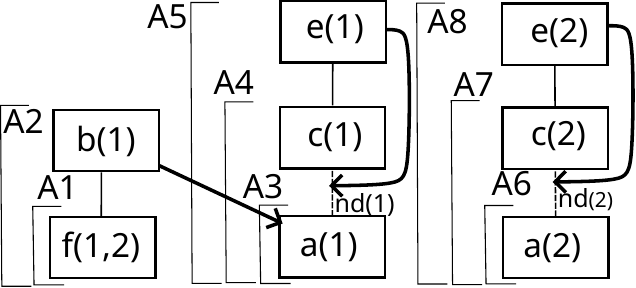}
        \caption{Argumentation theory}
        \label{fig:preliminars}
    \end{subfigure}
    \hspace{0.001\textwidth}  
    \begin{subfigure}[b]{0.22\textwidth}
        \centering
        \includegraphics[width=0.6\linewidth]{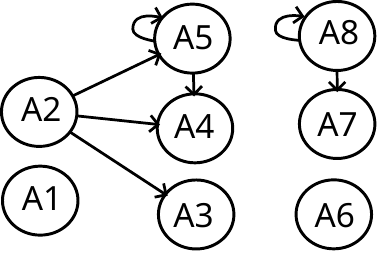}
        \caption{Argumentation graph}
        \label{fig:preliminars_aaf}
    \end{subfigure}
    \caption{Argumentation theory and induced AF from Example~\ref{ex:example_0}.}
    \label{fig:combined_preliminars}
\end{figure}

\begin{example}
Consider the argumentation theory $\asppv = \aspp$ with $\kbsa = \{ f(1,2) \}$\footnote{In this section we treat an expression like $f(1,2)$ as a propositional atom; later we define how such atoms are obtained via grounding of first-order atoms.}, $\kbso = \{ a(1), a(2) \}$, $\ruls= \{ \srul{f(1, 2)}{b(1)}, \srul{c(1)}{e(1)}, \srul{c(2)}{e(2)}\}$, $\ruld= \{ \drul{n_d(1)}{a(1)}{c(1)}, \drul{n_d(2)}{a(2)}{c(2)}\}$, and $\con=\{ \acrul{a(1)}{b(1)}, \acrul{a(2)}{b(2)},  \acrul{c(1)}{d(1)}, \acrul{c(2)}{d(2)}, \acrul{n_d(1)}{e(1)}, \acrul{n_d(2)}{e(2)} \}$. The set of arguments (A1-A8) of $\asppv$ and the attacks (arrows) between them is shown in Fig. \ref{fig:preliminars}. The \emph{AF} defined by $\asppv$ is depicted in Fig. \ref{fig:preliminars_aaf}. 
For this framework, the unique complete extension is $\{ A1, A2, A6 \}$, while the admissible sets are all of the subsets of $\{ A1, A2, A6 \}$. 
This means that the three arguments ($A1$, $A2$, $A6$) are all credulously accepted under the admissible semantics, whereas they are skeptically accepted under the complete semantics. In terms of claims, this implies that the conclusions ($f(1, 2), b(1), a(2)$) of the three arguments are all credulously (resp. skeptically) accepted under the admissible (resp. complete) semantics. Since the complete extension is unique in this case, the \emph{grounded} and \emph{preferred} extensions coincide with the complete extension. In contrast, there is no stable extension in this framework.
\label{ex:example_0}
\end{example}

\subsection{Datalog}
\label{subsec:Datalog}

To define Datalog (see, for instance,~\cite{abiteboul1995}) we extend the language of atoms $\lan$ to a language $\lam$ to be built from mutually disjoint, countably 
infinite sets of \emph{constants} $\cons$, \emph{variables} $\vars$, and \emph{predicates} $\preds$.  A \emph{term}
$\trmv$ is an element $\trmv \in \cons \cup \vars$. We represent a list of terms $\trmv_1,\ldots,\trmv_m$ as $\tlv$, and we 
treat such lists as sets, if appropriate.  An (extended) \emph{atom} is then an expression $\predv(\tlv)$ where $\predv \in \preds$. 
An atom is \emph{ground} if no variables occur in the atom (\ie~all terms are constants). 
For an expression $\phi$ (usually an atom or set of 
atoms), we denote by $\phi(\xlv)$ that $\phi$ uses (exactly) the variables $\xlv$. Throughout this work we will use upper-case letters for
variables and lower-case letters for constants.

\begin{definition}  A \emph{Datalog program} is a finite set of \emph{(Datalog) rules} of the form $\grul{\gbodv{\xlv}}{\ghedv{\ylv}}$ where $\ghedv{\ylv} \in \lam$ and $\gbodv{\xlv} \subseteq \lam$. The rule is \emph{safe} iff $\ylv \subseteq \xlv$.  
\end{definition}

Datalog programs are evaluated by grounding the rules and then computing their consequences:  

\begin{definition}\label{def:dat-grd} For a Datalog program $\datpv$, the \emph{Herbrand universe} $\heru{\datpv}$ is the 
set of all constants occurring in $\datpv$.  The \emph{Herbrand literal base} $\herl{\datpv}$ is the set of all ground atoms
 $\predv(\tlv)$ with $\predv$ occurring in $\datpv$ and $\tlv \subseteq \heru{\datpv}$.  The \emph{grounding of a rule} 
 $\grulv = \grul{\gbodv{\xlv}}{\ghedv{\ylv}} \in \datpv$ is defined as 
 $\ground{\grulv}{\datpv} = \{\grulv\sigma \mid \sigma: \xlv \cup \ylv \mapsto \heru{\datpv}\}$. 
 The \emph{grounding of the program} $\datpv$, $\groundp{\datpv}$, is the union of the grounding of all of its rules.    
\end{definition}

\begin{definition} 
The \emph{immediate consequence operator} $\ico{\datpv}$ for a Datalog program $\datpv$ and 
a set of ground atoms $I \subseteq \herl{\datpv}$ is $\icot{\datpv}{I} = \{\ghvs \mid \grul{\gbcv}{\ghvs} \in \groundp{\datpv}, 
\gbcv \subseteq I\}$. The least fixed-point of $\ico{\datpv}$ is $\icop{\datpv}{\infty} = \bigcup_{i \geq 0} \icop{\datpv}{i}$, 
where $\icop{\datpv}{0} = \emptyset$ and $\icop{\datpv}{i+1} = \icot{\datpv}{\icop{\datpv}{i}}$.  A ground atom  
$\atmv$ is \emph{derived} from $\datpv$ iff $\atmv \in \icop{\datpv}{\infty}$. A \emph{query} $\gqr{\datpv}{\qrv}$ asks 
for all atoms $\qrv(\tlv)$ that can be derived from $\datpv$.  
\end{definition}

Datalog can be extended to include a limited form of negation: 

\begin{definition}\label{def:datstrat} 
Datalog with \emph{stratified negation} extends Datalog 
by allowing negated 
atoms ${\sim}\predv(\tlv)$ in rule bodies of a Datalog program $\datpv$ as long as $\datpv$ has a \emph{stratification}.  
The latter is a function $l$ that assigns each predicate occurring in $\datpv$ a natural number such that for every rule
 $\rulv \in \datpv$ with $p \in \preds$ occurring in the head of $\rulv$, if $p' \in \preds$ occurs in the body of $\rulv$, 
 then $l(p) \geq \l(p')$, while if $p' \in \preds$ occurs in a negated atom in the body of $\rulv$, then $l(p) > \l(p')$.      
\end{definition}

\noindent Stratified Datalog programs are partitioned into strata based on predicate dependencies, with each stratum containing 
rules defining predicates at the same level. Evaluation proceeds bottom-up: at each stratum, rules are applied using facts derived 
so far—negated atoms are evaluated as true (and, thus, can be essentially deleted from rules) if no matching positive fact is derived 
in the prior lower stratum.

\section{Grounding rule-based argumentation}

\subsection{First-order rule-based argumentation in the ASPIC+ framework}
\label{subsec:foaspic}

Although a first-order (F.O.) syntax for rule-based ASPIC+ is often used, it has, to our knowledge, not been formally defined.  We provide one following common definitions of logic-programming (see e.g.~\cite{Faber20}). For this we again make use of a language $\lam$ of atoms as in Section~\ref{subsec:Datalog}, while $\lan \subseteq \lam$ denotes the ground atoms. 

\begin{definition}\label{def:foaspic} A \emph{(first-order) \argpr} is a tuple $\asppv = \aspp$. $\con$ is a finite set of (safe) 
\emph{contrary expressions} $\acrul{\astmv{\xlv}}{\astmsv{\xlv}}$
    with $\astmv{\xlv} \in \lam$, $\astmsv{\xlv} \subseteq \lam$.  Moreover, no constants occur in either 
    $\astmv{\xlv}$ or $\astmsv{\xlv}$.  $\ruls$ is a finite set of (safe) \emph{strict} rules, having the form 
    $\srul{\abodv{\xlv}}{\ahedv{\ylv}}$ with $\ahedv{\ylv} \in \lam$, $\abodv{\xlv} \subseteq \lam$, $\ylv \subseteq \xlv$. $\ruld$ 
    is a finite set of (safe) \emph{defeasible} rules, which have the form $\drul{\anamv{\zlv}}{\abodv{\xlv}}{\ahedv{\ylv}}$ 
    with $\anamv{\zlv} \in \lam$, $\ahedv{\ylv} \in \lam$, $\abodv{\xlv} \subseteq \lam$, $\ylv \subseteq \xlv$,
    $\zlv \subseteq \xlv$.  Finally, $\kbsa \subseteq \lan$ and $\kbso \subseteq \lan$, for which $\kbsa \cap \kbso = \emptyset$, 
    are the set of \emph{facts} and \emph{assumptions} respectively. Both $\kbsa$ and $\kbso$ are also finite. 
\end{definition}

\begin{example} An example of a first-order \argpr~is $\asppv = \aspp$ where $\con=\{ \acrul{a(X)}{b(X)}, \acrul{n_d(X)}{e(X)},  \acrul{c(X)}{d(X)} \}$, $\ruls= \{ \srul{f(X, Y)}{b(X)}, \srul{c(X)}{e(X)}\}$, $\ruld= \{ \drul{n_d(X)}{a(X)}{c(X)} \}$, $\kbso=\{ a(1), a(2) \}$, $\kbsa = \{ f(1,2) \}$. As will be shown in the following sections, the grounding of this first-order theory is equivalent, in terms of induced AFs, to the propositional theory of Example~ \ref{ex:example_0} depicted in Fig~\ref{fig:preliminars}.
\label{ex:example_1}
\end{example}
 
As to the restrictions in Definition~\ref{def:foaspic}, safety of rules is a common restriction from logic-programming to ensure termination when solving.  The restriction of facts and assumptions to be ground is to distinguish these from strict rules and defeasible rules (\ie~ if non-ground assumptions, for instance, are needed, they can be defined via defeasible rules).  As to the restriction for contrary expressions, this is for simplicity; if a contrary expression like $\acrul{p(X)}{q(X, Y, c)}$ is needed, it can be defined via the contrary expression $\acrul{p(X)}{q'(X)}$ and the rule $\srul{q(X, Y, c)}{q'(X)}$.

To obtain a ground (\ie~propositional) theory from a F.O. argumentation theory the same approach as for Datalog 
(Section~\ref{subsec:Datalog}) can be used.  I.e. for a theory $\asppv$, the Herbrand universe $\aheru{\asppv}$ is 
the set of all constants occurring in $\asppv$.  Then, the grounding of $\asppv$, $\agroundp{\asppv}$, is obtained 
by grounding each of the contrary expressions and rules in $\asppv$, \ie~replacing all variables for elements of 
$\aheru{\asppv}$ in all possible ways.  The semantics of $\asppv$ is then obtained by evaluating $\agroundp{\asppv}$ 
 via the induced AF as explained in Section~\ref{subsec-prop-aspic}.

\subsection{Grounding via Datalog: the basics}
\label{subsec:grd1}

Naively grounding argumentation theories as described at the end of Section~\ref{subsec:foaspic} will often produce rules that do not form part of any argument. 

\begin{example} Consider the argumentation theory $\asppv$ of Example~\ref{ex:example_1} and let us focus on the strict rule $\srul{f(X, Y)}{b(X)}$. A naive grounding produces $\agroundp{\ruls} = \{ \srul{f(1, 1)}{b(1)}, \srul{f(1, 2)}{b(1)}, \srul{f(2, 1)}{b(2)}, \srul{f(2, 2)}{b(2)}\}$.  
The only arguments derived by these rules and included in the AF for $\asppv$ use $f(1,2)$ as a premise and $b(1)$ as a conclusion, via the rule $\srul{f(1, 2)}{b(1)}$ (\ie~the arguments A1 and A2 depicted in Fig~\ref{fig:preliminars}). The same set of arguments would be produced with a more efficient grounding that only produces the rule $\srul{f(1, 2)}{b(1)}$. 
\label{ex:example_1_naive_gr}
\end{example}
 
\noindent We thus now introduce a transformation of an ASPIC+ theory $\asppv$ that generates a set of Datalog rules, which helps
 to produce a smaller grounding that excludes unnecessary rules, while preserving the extensions of  $\asppv$.     

\begin{transformation} Let $\asppv = \aspp$ be an \argpr. For a rule $\rulv = \srul{\abodv{\xlv}}{\ahedv{\ylv}} \in \ruls$ or $\rulv = \drul{\anamv{\zlv}}{\abodv{\xlv}}{\ahedv{\ylv}} \in \ruld$ 
we denote the transformation into Datalog rules by $\hat{r}$ as the set of rules consisting of the following:

\begin{align}
\grul{\gbodv{\xlv}}{&\aux{\xlv}} \label{eq:transf1_1} \\
\grul{\aux{\xlv}}{&\ghedv{\ylv}} \label{eq:transf1_2}
\intertext{where $\auxn{\rulv}$ is a fresh predicate that is unique for the rule.
For the defeasible rule $\rulv \in \ruld$ we also need the following additional Datalog rule:}
\grul{\aux{\xlv}}{&\anamv{\zlv}} \label{eq:transf1_3}
\end{align}
\label{def:transf1}
\noindent Facts and assumptions are transformed into Datalog rules with empty bodies. In particular, for any ground atom $b\in \kbsa \cup \kbso$ we define $\hat{b}= \grul{}{b}$.  Then, the transformation of $\asppv$ to the respective Datalog program is 
$\datpv_\asppv = \{\hat{\ruls} \cup \hat{\ruld} \cup \hat{\kbsa} \cup \hat{\kbso}\}$.
\end{transformation}

Once we have created the Datalog program $\datpv_\asppv$ by applying Transformation~\ref{def:transf1} for a theory $\asppv$, we ground $\asppv$ 
by querying a Datalog engine for each strict and defeasible rule $r \in \rul$ with the respective query $(\datpv_\asppv, \auxn{\rulv})$ for the introduced
auxiliary predicates. For all obtained ground atoms $\aux{\vec{a}_i}$, with $0\leq i \leq l$, we make the ground substitutions 
$\{r\sigma_i \mid \sigma_i: \xlv \mapsto \vec{a}_i\}$ as described in Algorithm \ref{alg:ground_tranf1}. The union over all ground substitutions of a rule $r$ for the Datalog program $\datpv_\asppv$ is defined as
 $\grounddl{\rulv}{\datpv_\asppv} = \bigcup_{i=0}^l r\sigma_i$.

\begin{algorithm}
\caption{Grounding of an ASPIC+ rule}
\begin{algorithmic}[1]\label{alg:ground_tranf1}
\REQUIRE A theory $\asppv$, $\rulv \in \rul$, $\datpv_\asppv$ with $n_r(\xlv)\in\datpv_\asppv$
\ENSURE $\grounddl{\rulv}{\datpv_\asppv}$
\STATE{$\grounddl{\rulv}{\datpv_\asppv} \longleftarrow \emptyset$}
\FORALL {$0\leq i \leq l$ answers $\aux{\vec{a}_i}$ to $\gqr{\datpv_\asppv}{\auxn{\rulv}}$ }
    \STATE {// each $\aux{\vec{a_i}}$ gives rise to a ground instance of $\rulv$}
	  \STATE { $\{  \sigma_i : X_k \mapsto a_k \mid 1 \leq k \leq |\xlv| \}$}
		\STATE {$\grounddl{\rulv}{\datpv_\asppv} \longleftarrow \grounddl{\rulv}{\datpv} \cup \{r\sigma_i\}$}
\ENDFOR
\RETURN $\grounddl{\rulv}{\datpv_\asppv}$
\end{algorithmic}
\label{alg:ground_tranf1}
\end{algorithm}

\noindent Grounding of a contrary expression $c: \acrul{\astmv{\xlv}}{\astmsv{\xlv}}$ is obtained in an analogous way (thus, we will often say that we also ground a contrary expression following Algorithm~\ref{alg:ground_tranf1}). 
We query the Datalog engine with the query $(\datpv_\asppv, s)$ and for each answer $s(\vec{a}_i)$ with $0\leq i \leq l$ we
make the substitution $\{c\sigma_i \mid  \sigma_i : X_k \mapsto a_k \mid 1 \leq k \leq |\xlv| \}$.
Then, the union over all ground substitutions of a contrary expression $c$ for the Datalog program $\datpv_\asppv$ is defined as
$\grounddl{c}{\datpv_\asppv} = \bigcup_{i=0}^l c\sigma_i$.  For $\asppv = \aspp$ we thus obtain the grounding $\grounddlp{\asppv}= (\grounddlp{\con}, \grounddlp{\ruls},\grounddlp{\ruld}, \kbsa, \kbso)$, 
with $\grounddlp{\con} = \bigcup_{c\in  \con} \grounddl{c}{\datpv_\asppv}$ and analogously for $\grounddlp{\ruls}$ and $\grounddlp{\ruld}$.   

\begin{example} In this example we show how a propositional argumentation theory $\grounddlp{\asppv}$ is obtained by 
grounding the F.O. theory from Example~\ref{ex:example_1} using the Datalog program $\datpv_\asppv$ and 
Algorithm~\ref{alg:ground_tranf1}. 
To obtain $\datpv_\asppv$, we first transform $f(1,2) \in \kbsa$, $a(1) \in \kbso$ and $a(2) \in \kbso$ into the Datalog rules $\grul{}{f(1,2)}$, $\grul{}{a(1)}$ and $\grul{}{a(2)}$, and add them to 
$\datpv_\asppv$. Next, we apply Transformation~\ref{def:transf1} to strict and defeasible rules. For instance, 
$\srul{f(X, Y)}{b(X)}$ results in the Datalog rules $\grul{f(X, Y)}{\aux{X,Y}}$ and $\grul{\aux{X,Y}}{b(X)}$, 
both of which are added to $\datpv_\asppv$. Similarly, the 
defeasible rule $\drul{n_d(X)}{a(X)}{c(X)}$ results in $\grul{a(X)}{\auxp{X}}$ and $\grul{\auxp{X}}{c(X)}$,  
along with the additional rule $\grul{\auxp{X}}{n_d(X)}$. Now, we ground 
$\asppv$ by invoking Algorithm~\ref{alg:ground_tranf1} for each rule of the theory.
For example, consider the rule  $\srul{f(X, Y)}{b(X)}$. First, we execute the query $(\datpv_\asppv, \auxn{\rulv})$, 
producing the ground atom $\aux{1,2}$ and the substitution $\{  \sigma : X \mapsto 1, Y \mapsto 2 \}$. Then, we apply 
this substitution to the rule obtaining $\srul{f(1, 2)}{b(1)}$. By applying the same procedure to the other rules ($\srul{c(X)}{e(X)}$ and $\drul{n_d(X)}{a(X)}{c(X)}$) we obtain: $\srul{c(1)}{e(1)}$; $\srul{c(2)}{e(2)}$; $\drul{n_d(1)}{a(1)}{c(1)}$ and $\drul{n_d(2)}{a(2)}{c(2)}$. Now, let us  consider the contrary relation $\acrul{a(X)}{b(X)}$. We execute the query $(\datpv_\asppv, a)$ obtaining the substitutions $\{  \sigma_1 : X \mapsto 1 \}$ and $\{  \sigma_2 : X \mapsto 2 \}$, resulting in the propositional contrary relations  $\acrul{a(1)}{b(1)}$ and $\acrul{a(2)}{b(2)}$.  
The obtained propositional rules and contrary relations, together with the initial facts and 
assumptions ($\kbsa$ and $\kbso$), forms the grounded argumentation theory $\grounddlp{\asppv}$, which, as anticipated, coincides with the propositional argumentation theory in Example~\ref{ex:example_0}.
\label{ex:trasl_1}
\end{example}

Lemma~\ref{lem:grd1} expresses that the AFs defined by $\agroundp{\asppv}$ and $\grounddlp{\asppv}$ are the same, from which it clearly follows (Theorem~\ref{thm:grd1}), that $\asppv$ and $\grounddlp{\asppv}$ have the same extensions.  

\begin{lemma}[$\star$\footnote{Statements marked by ``$\star$'' are proven in the appendix.} ]\label{lem:grd1} 
Let $\asppv$ be an \argth~and $\grounddlp{\asppv}$ the grounding via the Datalog program $\datpva$ as per Transformation~\ref{def:transf1}. Then, $\arts{\agroundp{\asppv}} = \arts{\grounddlp{\asppv}}$ and 
$\atts{\agroundp{\asppv}} = \atts{\grounddlp{\asppv}}$.
\end{lemma}

\begin{theorem}[$\star$]\label{thm:grd1} Let $\asppv$ be an \argth~and $\grounddlp{\asppv}$ the grounding via the Datalog program $\datpva$ as per Transformation~\ref{def:transf1}. Then, $\semv(\asptv)$ $=$ $\semv(\agroundp{\asppv})$ $=$ $\semv(\grounddlp{\asppv})$ for $\semv \in \semlist$.
\end{theorem}

\subsection{Improvements}

\subsubsection{Non-approximated predicates.}\label{subsec:simp1} In Transformation~\ref{def:transf1} we ignore the distinction between defeasible and non-defeasible elements (assumptions and defeasible rules on the one side, facts and strict rules on the other).  To obtain all possible arguments of an argumentation theory this is fine.  But we can further simplify the grounding by generating rules that will be used only in acceptable arguments (\ie~those included in some extension).  

\begin{example}
Consider the propositional theory obtained in Example~\ref{ex:trasl_1} (via grounding the F.O. theory from Example~\ref{ex:example_1} 
via the procedure described in Section~\ref{subsec:grd1}) and depicted in Figure~\ref{fig:preliminars}. Any argument using $a(1)$ as a 
premise (namely $A3$, $A4$, and $A5$)  is undermined by $A2$, which is itself not attacked by any argument. As a result, neither 
$A3$, $A4$ nor $A5$ will be included in any extension for any of the semantics we consider in this work.  
Hence, the assumption and rules that generate such arguments (\ie~$a(1)$, $\drul{n_d(1)}{a(1)}{c(1)}$, and $\srul{c(1)}{e(1)}$) 
can also be excluded from the grounding, further reducing the size of the grounding while preserving the extensions. 
\end{example}

We introduce the notion of approximated and non-approximated predicates to help us formalise when rules and assumptions can be removed from the grounding in Definition~\ref{def:approx}.  The intuition behind the distinction between these different predicates is as follows: non-approximated predicates are those for which grounding can fully determine derivability of the ground instances within some extension. Approximated predicates leave determination of this to the solver.

\begin{definition}\label{def:approx} Let $\asppv = \aspp$ be an \argth~and let~$\predv$ first be a predicate and $\rulv \in \rul$ a rule whose head is an atom containing $\predv$.  Then, predicate $\predv$ \emph{depends positively} on a predicate $\predv$' if the body of $\rulv$ contains an atom containing $\predv$'. The predicate $\predv$ \emph{depends negatively} on $\predv'$ if  $\predv'$ appears in the contrary expression of one of the defeasible elements of $\rulv$ ($\cona{\defe{\rulv}}$).  Let now $\predv$ be a predicate occurring in an assumption atom $l$ ($l \in \kbso$).  Then, $\predv$ \emph{depends negatively} on $\predv'$ if $\predv'$ occurs in the contrary expression of $l$. The set of \emph{approximated} predicates is the minimal set containing a predicate $\predv$ if one of the following cases holds: i) $\predv$ depends on an approximated predicate, or ii) there is a circular sequence of dependencies $\predv = \predv_1,\predv_2, \ldots, \predv_n = \predv$, where each $\predv_{i+1}$ depends on $\predv_i$ and there is a $\predv_{i+1}$ that depends negatively  on $\predv_i$.  Any predicate that is not approximated we call \emph{non-approximated}.  
\end{definition}

\begin{example} Consider the argumentation theory of Example~\ref{ex:example_1}. To compute the approximated and non-approximated predicates of the theory, we analyze the positive and negative dependencies between predicates. For example, consider the rules $r_1 = \drul{n_d(X)}{a(X)}{c(X)}$ and $r_2 = \srul{c(X)}{e(X)}$, and the contrary expression $\acrul{n_d(X)}{e(X)}$. We consider the predicates appearing in the heads of the rules, namely $c$ and $e$. First, from the rule $r_2$, we see that $e$ depends positively on $c$. Next, rule $r_1$ shows that $c$ depends negatively on $e$. This is because $e$ appears in the contrary expression of $n_d(X)$, which is a defeasible element of $r_1$. Therefore, there exist circular sequences of dependencies with at least one negative relation: $e,c,e$ and $c,e,c$.  This indicates that both $e$ and $c$ are approximated predicates. In contrast, all other predicates in the theory are non-approximated predicates.
\label{ex:appr_pred}
\end{example}

Using the knowledge of which predicates are non-approximated, we can simplify the grounding:

\begin{transformation}\label{def:transf2} Let $\asppv = \aspp$ be an \argth. A rule $\rulv = \srul{\abodv{\xlv}}{\ahedv{\ylv}} \in \ruls$ or $\rulv = \drul{\anamv{\xlv}}{\abodv{\ylv}}{\ahedv{\zlv}} \in \ruld$, given $\{ l_1(\vec{X_1}), \ldots , l_n(\vec{X_n}) \} \subseteq \cona{\defe{\rulv}}$ where the $l_i$ are exactly the  non-approximated predicates in $\cona{\defe{\rulv}}$, gives rise to the set $\hat{\rulv}$ consisting of the following Datalog rules:
 \begingroup   
 \setcounter{equation}{0}
\begin{align}
    \grul{\gbodv{\xlv}, \sim l_1(\vec{X_1}), \ldots , \sim l_n(\vec{X_n})}{&\aux{\xlv}}    \label{eq:transf2_1} \\
    \grul{\aux{\xlv}}{&\ghedv{\ylv}}    \label{eq:transf2_2}
    \intertext{where $\auxn{\rulv}$ is a fresh predicate that is unique for the rule.
    For the defeasible rule $\rulv \in \ruld$ the following rule also is part of $\hat{\rulv}$ as in Transformation~\ref{def:transf1}:}
\grul{\aux{\xlv}}{&\anamv{\zlv}} \label{eq:transf2_3}
\end{align}
\endgroup

Facts are also transformed as in Transformation~\ref{def:transf1}: for any $b\in \kbsa$, $\hat{b}= \grul{}{b}$. Assumptions are transformed analogously to defeasible rules.  I.e. for a ground atom $b\in\kbso$ with $\{ l_1(\tlv_1)), \ldots , l_n(\tlv_n) \} \subseteq \cona{b}$ where $l_i$ are exactly the non-approximated predicates appearing in $\cona{b}$, we define $\hat{b} =  \grul{ {\sim}l_1(\tlv_1), \ldots , {\sim}l_n(\tlv_n)}{b}$. Then, the transformation of the argumentation theory $\asppv$ to the respective Datalog program is
$\datpv_\asppv =\{\hat{\ruls} \cup \hat{\ruld} \cup \hat{\kbsa} \cup \hat{\kbso}\}$.  
\end{transformation}

\noindent We observe that, because only non-approximated predicates appear negated in bodies of rules of $\datpv_\asppv$ as per Transformation~\ref{def:transf2}, $\datpv_\asppv$ 
uses stratified negation as defined in Section~\ref{subsec:Datalog}.  
The grounding of $\asppv$ via $\datpv_\asppv$ as per Transformation~\ref{def:transf2} is then obtained by grounding the contrary relation and rules of $\asppv$ by querying $\datpv_\asppv$ as in Section~\ref{subsec:grd1}.  The difference is that now also assumptions are queried, \ie~$\grounddlp{\kbso} = \{b(\tlv) \in \kbso \cap \gqr{\datpv_\asppv}{b}\}$.  For $\asppv = \aspp$ we thus obtain the grounding $\grounddlp{\asppv}= (\grounddlp{\con}, \grounddlp{\ruls},\grounddlp{\ruld}, \kbsa, \grounddlp{\kbso})$\footnote{We use the same notation, \ie~ $\datpv_\asppv$ and $\grounddlp{\asppv}$, for the different versions of the Datalog program and groundings we introduce.}.

\begin{figure}[h]
    \centering
    \begin{subfigure}[b]{0.22\textwidth}
        \centering
        \includegraphics[width=\linewidth]{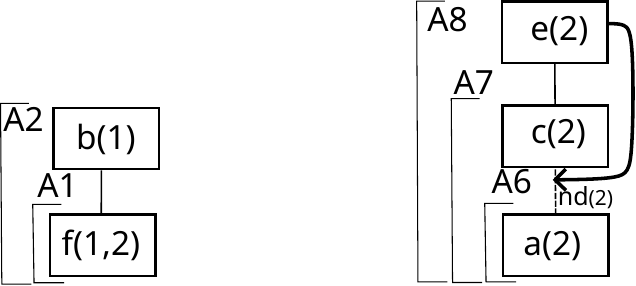}
        \caption{Argumentation theory}
        \label{fig:transf_1}
    \end{subfigure}
    \hspace{0.001\textwidth}  
    \begin{subfigure}[b]{0.22\textwidth}
        \centering
        \includegraphics[width=0.6\linewidth]{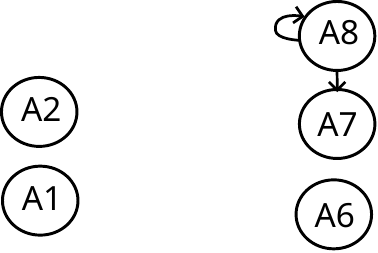}
        \caption{Argumentation graph}
        \label{fig:transf_1_aaf}
    \end{subfigure}
    \caption{Argumentation theory and induced AF from Example~\ref{ex:example_2}.}
    \label{fig:combined_transf_1}
\end{figure}

\begin{example}\label{ex:example_2} We ground the argumentation theory from Example~\ref{ex:example_1} using the Datalog program $\datpv_\asppv$ obtained via Transformation~\ref{def:transf2}. As to the facts and strict rules, Transformation~\ref{def:transf2} is exactly as Transformation~\ref{def:transf1}, \ie~as in Example~\ref{ex:trasl_1} (since, in particular, $\defe{\rulv} = \emptyset$ for $\rulv \in \ruls$).  Thus, we focus on the transformation of defeasible rules and assumptions. Consider the defeasible rule $\drul{n_d(X)}{a(X)}{c(X)}$, which includes two defeasible elements ($c(X)$ and $n_d(X)$), whose contraries are $e(X)$ and $d(X)$. As discussed in Example~\ref{ex:appr_pred}, $e$ is an approximated predicate, and therefore it does not appear in the Datalog rules generated by Transformation~\ref{def:transf2}. The resulting rules are  $\grul{a(X), {\sim}d(X)}{\auxp{X}}$ and $\grul{\auxp{X}}{c(X)}$,  along with the additional rule $\grul{\auxp{X}}{n_d(X)}$.  The assumptions $a(1)$ and $a(2)$ are translated into $\grul{ {\sim}b(1)}{a(1)}$ and $\grul{ {\sim}b(2)}{a(2)}$ respectively, since the contrary of $a(1)$ (resp. $a(2)$) is $b(1)$ (resp. $b(2)$) and $b$ is a non-approximated predicate.

Next, we ground $\asppv$ by invoking Algorithm~\ref{alg:ground_tranf1} for each rule of the theory, as already shown in Example~\ref{ex:trasl_1}. Notably, we obtain fewer propositional rules compared with the previous example. For instance, consider again the rule $\drul{n_d(X)}{a(X)}{c(X)}$, which we ground by performing the query  $(\datpv_\asppv, \auxnp{\rulv})$ and obtaining the single 
substitution $\{  \sigma_1 : X \mapsto 1 \}$, instead of the two substitutions obtained in Example~\ref{ex:trasl_1}.  In fact, the second substitution $\{  \sigma_2 : X \mapsto 2 \}$ can not be derived due to the negative terms introduced by Transformation~\ref{def:transf2}. The grounding of contrary relations is also impacted by the introduced optimization. For example, grounding the contrary relation $\acrul{a(X)}{b(X)}$ now results in a single propositional expression, $\acrul{a(2)}{b(2)}$, instead of the two previously derived.
On the other hand, $a(1) \not\in \grounddlp{\kbso}$ as $a(1) \not\in \gqr{\datpv_\asppv}{a}$. Finally, the obtained assumptions, rules and contrary relations, together with the initial facts, forms the grounded argumentation theory $\grounddlp{\asppv}$, depicted in Fig~\ref{fig:combined_transf_1} together with the induced AF. Note that, although the arguments A3, A4, A5 from the theory of Example~\ref{ex:example_0} are lost, the unique complete (as well as grounded and preferred) extension is still $\{ A1, A2, A6 \}$ (while there  is also no stable extension) and, thus, also the acceptable conclusions are the same as in Example~\ref{ex:example_0}. In this example the admissible extensions are also the same but we show in Example~\ref{ex:counter} that this is not guaranteed when grounding via Transformation~\ref{def:transf2}.  
\end{example}

Lemma~\ref{lem:grd2} expresses the relation between the AFs induced by $\agroundp{\asppv}$ and the improved $\grounddlp{\asppv}$.  For this, for an argumentation theory, we define the \emph{certain} arguments to be those that are included in every complete extension of the theory.  If, on the other hand, an argument is not attacked by a certain argument we call it \emph{tentative}. For a set of arguments $\argsv$, we then denote $\artsa{\argsv}$ to be the certain arguments in $\argsv$, while $\artse{\argsv}$ are the tentative arguments in $\argsv$.  Note, in particular, that certain arguments are also tentative. We also define $\artpr{\argsv}{P}$ to be those arguments in $\argsv$ with conclusions making use of predicates in a set of predicates $P$. Similarly, $\aproj{\atts{\agroundp{\asppv}}}{\arts{\grounddlp{\asppv}}} = \{(\argv,\argv') \in \atts{\agroundp{\asppv}} \mid \{\argv,\argv'\} \subseteq \arts{\grounddlp{\asppv}}\}$.  Then, Lemma~\ref{lem:grd2} indicates that the AFs induced by $\agroundp{\asppv}$ and $\grounddlp{\asppv}$ coincide on the certain arguments.  That these are the arguments that count for all semantics that produce complete extensions is expressed in Theorem~\ref{thm:grd2}.

\begin{lemma}[$\star$]\label{lem:grd2} Let $\asppv$ be an \argth~and $\grounddlp{\asppv}$ the grounding via the Datalog program $\datpva$ as per Transformation~\ref{def:transf2}. Then, $\artse{\agroundp{\asppv}} \subseteq \arts{\grounddlp{\asppv}} \subseteq \arts{\agroundp{\asppv}}$ and $\atts{\grounddlp{\asppv}} = \aproj{\atts{\agroundp{\asppv}}}{\arts{\grounddlp{\asppv}}} $.
\end{lemma}

\begin{theorem}[$\star$]\label{thm:grd2} Let $\asppv$ be an \argth~and $\grounddlp{\asppv}$ the grounding via the Datalog program $\datpva$ as per Transformation~\ref{def:transf2}. Then, $\semv(\asptv)$ $=$ $\semv(\agroundp{\asppv})$ $=$ $\semv(\grounddlp{\asppv})$ for $\semv \in \semlistc$.
\end{theorem}

\begin{example}\label{ex:counter} Note, on the other hand, that it is not the case that $\adm(\agroundp{\asppv})$ $=$ $\adm(\grounddlp{\asppv})$. This can be seen by considering the simple (ground) argumentation theory $\asppv$ with $\kbsa = \{a\}$, $\kbso = \{b,c\}$, $\rul = \emptyset$, and $\con = \{\crul{b}{a}, \crul{c}{b}\}$.  Then,  $\agroundp{\asppv} = \asppv$. On the other hand, $\datpva = \{\grul{}{a}, \grul{{\sim}a}{b},\grul{{\sim}b}{c} \}$ and, hence,  $\grounddlp{\asppv}$ is formed by $\kbsa' = \kbsa = \{a\}$, $\rul'  = \rul = \emptyset$, while $\kbso' = \{c\}$ and $\con' = \{\crul{c}{b}\}$ (but $b \not\in \kbso'$).  Then, $\adm(\agroundp{\asppv}) = \{\emptyset,\{a\},\{a,c\}\}$, while $\adm(\grounddlp{\asppv}) = \adm(\agroundp{\asppv}) \cup \{\{c\}\}$.      
\end{example}

\subsubsection{Facts.} In the previous section we simplified the grounding by excluding rules appearing only in arguments that do not appear in any extension.  We now look at what is, to some extent, the dual view.

\begin{example} Consider the propositional theory obtained in Example~\ref{ex:example_2} (via grounding the F.O. theory from Example~\ref{ex:example_1} using the procedure described in the previous section) and depicted in Figure~\ref{fig:combined_transf_1}.
Consider the fact $f(1,2)$ and the strict rule $\srul{f(1,2)}{b(1)}$ of Example~\ref{ex:example_1}, which together make up the argument A2. This argument, as it is made up only of facts and strict rules, is immune to attacks, which means that the argument A2 appears in all extensions for all the semantics we consider in this work.  This means also that the grounding can be simplified by adding the claim $b(1)$ to the fact base and, as a consequence, removing the strict rule from the set of rules as it becomes redundant.     
\label{ex:example_4}
\end{example}

Towards simplifying the grounding along the lines of the idea explored in Example~\ref{ex:example_4}, note that the dependency relations between predicates of an argumentation theory from Definition~\ref{def:approx} can be extended to rules. In particular, for the positive dependency relation: a rule $\rulv$ depends positively on $\rulv'$ if the predicate $p$ appearing in the head of $\rulv$ depends positively on the predicate $p'$ occurring in the head of $\rulv'$ (i.e. $p'$ occurs in the body of $\rulv$ as well as the head of $\rulv'$).  The positive dependency graph $\depgp$ for the set of rules $\rul$ of an argumentation theory $\asppv$ is then a directed graph having rules as nodes and an edge from $\rulv'$ to $\rulv$ indicating that $\rulv$ depends positively on $\rulv'$.  Given the dependency graph $\depgp$, the strongly connected components of $\depgp$ form a topological ordered partition into sub-graphs $L_{\asppv} = (C_1, C_2, \dots, C_n)$.  Algorithm~\ref{alg:ground_theory} then improves on the previous groundings by also (in lines 3-18) iterating over the strongly connected components of $\depgp$ and collecting facts, while removing strict rules, as illustrated in Example~\ref{ex:example_4}\footnote{In the algorithm $B(r)$ denotes the body of a rule $r$, and $h(r)$ the head.}.

\begin{algorithm}[h]
\caption{Grounding of an ASPIC+ theory}
\label{alg:ground_theory}
\begin{algorithmic}[1]
\REQUIRE A theory $\asppv = \aspp$, $L_\asppv$, the Datalog program $\datpva$ obtained via Transformation~\ref{def:transf2}.
\ENSURE $\grounddlp{\asppv}$.

\STATE $\grounddlp{\kbsa} \leftarrow \kbsa$
\STATE $\grounddlp{\rul} \leftarrow \emptyset$

\FORALL{$C \in L_\asppv$}
\STATE { // Ground rules in $C$ via Algorithm~\ref{alg:ground_tranf1}.} \label{alg:line_call_alg_1}
\STATE $\grounddlp{C} \leftarrow \bigcup_{\rulv\in  C} \grounddl{\rulv}{\datpv_\asppv}$ 
    \REPEAT
    \STATE $\grounddlp{\kbsa}' \leftarrow \grounddlp{\kbsa}$
        \FORALL{$r \in \grounddlp{C}$}
            \STATE { // Delete facts from bodies of rules.}\label{alg:line_del_facts}
            \IF{$r \in \grounddlp{\ruls}$, $B(r) \setminus \grounddlp{\kbsa} = \emptyset$, and $h(r) \not\in \kbso$}
                \STATE $\grounddlp{\kbsa} \leftarrow \grounddlp{\kbsa} \cup \{ h(r) \}$ \label{alg:line_add_facts}
                \STATE $\grounddlp{C} \leftarrow \grounddlp{C} \setminus \{ r \}$        \label{alg:del_rule}
            \ENDIF
        \ENDFOR
    \STATE { // Repeat until no new facts are produced.}
    \UNTIL{$\grounddlp{\kbsa} = \grounddlp{\kbsa}'$} 
   \STATE $\grounddlp{\rul} \leftarrow \grounddlp{\rul} \cup \grounddlp{C}$
\ENDFOR
\STATE { // Ground assumptions and contraries as before.}
\STATE $\grounddlp{\kbso} \leftarrow \{b(\tlv) \in \kbso \cap \gqr{\datpva}{b}\}$
\STATE $\grounddlp{\con} \leftarrow \bigcup_{c\in  \con} \grounddl{c}{\datpv_\asppv}$ 
\STATE { // Form the grounded theory.}
\STATE $\grounddlp{\asppv} \leftarrow (\grounddlp{\con}, \grounddlp{\ruls},$
\STATE \hspace{5.9em} $\grounddlp{\ruld}, \grounddlp{\kbsa}, \grounddlp{\kbso})$
\RETURN $\grounddlp{\asppv}  $
\end{algorithmic}
\end{algorithm}

\begin{figure}[h]
        \centering
        \begin{subfigure}[b]{0.22\textwidth}
            \centering
            \includegraphics[width=\linewidth]{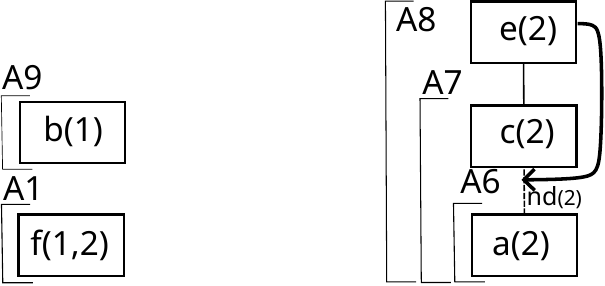}
            \caption{Argumentation theory}
            \label{fig:transf_1}
        \end{subfigure}
        \hspace{0.001\textwidth}  
        \begin{subfigure}[b]{0.22\textwidth}
            \centering
            \includegraphics[width=0.6\linewidth]{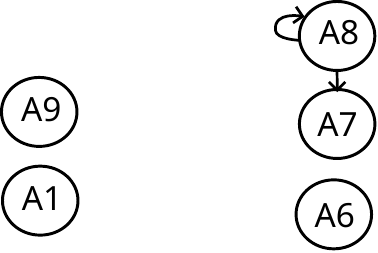}
            \caption{Argumentation graph}
            \label{fig:transf_1_aaf}
        \end{subfigure}
        \caption{Argumentation theory and induced AF from Example~\ref{ex:example_5}.}
        \label{fig:combined_transf_3}
    \end{figure}

\begin{example}\label{ex:example_5}

We ground the argumentation theory from Example~\ref{ex:example_1} using the Datalog program $\datpv_\asppv$ obtained via Transformation~\ref{def:transf2} and Algorithm~\ref{alg:ground_theory}, which incorporates the optimization described above.  Since this optimization does not affect assumptions or defeasible rules,  we only focus on the strict rules. Consider the rule $\srul{f(X,Y)}{b(X)}$. At line~\ref{alg:line_call_alg_1} of Algorithm~\ref{alg:ground_theory}, we call Algorithm~\ref{alg:ground_tranf1}, which generates the propositional rule $\srul{f(1,2)}{b(1)}$. Then, at line~\ref{alg:line_del_facts}, we remove the fact $f(1,2)$ from the body of the rule, resulting in the rule $\srul{}{b(1)}$. Because this is now a strict rule with an empty body, at lines~\ref{alg:line_add_facts}--\ref{alg:del_rule} we add $b(1)$ to $\grounddlp{\kbsa}$ and remove $\srul{}{b(1)}$ from $\grounddlp{\ruls}$. The resulting argumentation theory and argumentation graph is depicted in Figure~ \ref{fig:combined_transf_3}. In this case, the unique complete extension is $\{A1, A2, A9\}$, that differs from the one obtained in Example~ \ref{ex:example_1}. It is worth noting that, although extensions are not preserved, when considering the conclusions of the arguments in the extensions of both examples these coincide. In fact, the conclusions ($f(1, 2)$, $b(1)$, and $a(2)$) are skeptically accepted under the complete semantics, just as in the previous example.
    \end{example}

Lemma~\ref{lem:grd3}, and as a consequence Theorem~\ref{thm:grd3}, follow from the fact that the groundings obtained in the previous section and the one obtained via Algorithm~\ref{alg:ground_theory} differ only on the simplifications on the fact base and strict rules, from whence the AFs induced by both groundings are equivalent when considering acceptable claims.  The equivalence is not w.r.t. arguments since, as we have seen in Example~\ref{ex:example_5}, adding new facts and removing strict rules in the grounding via Algorithm~\ref{alg:ground_theory} clearly affects the arguments that can be constructed.

\begin{lemma}[$\star$] \label{lem:grd3} Let $\asppv$ be an \argth, $\grounddlph{\asppv}$ the grounding of $\asppv$ via Algorithm~\ref{alg:ground_theory}, and $\grounddlpt{\asppv}$ the grounding via the Datalog program $\datpva$ as per Transformation~\ref{def:transf2}\footnote{\ie~without the further simplifications in Algorithm~\ref{alg:ground_theory}.}. Then, $\{\conc{E} \mid E \in \semv(\grounddlpt{\asppv})\}$ $=$ $\{\conc{E} \mid E \in \semv(\grounddlph{\asppv})\}$ for $\semv \in \semlistc$.
\end{lemma}

\begin{theorem}[$\star$]\label{thm:grd3}\sloppy Let $\asppv$ be an \argth, $\grounddlph{\asppv}$ the grounding of $\asppv$ via Algorithm~\ref{alg:ground_theory}. Then, $\{\conc{E} \mid E \in \semv(\asptv)\}$ $=$ $\{\conc{E} \mid E \in \semv(\agroundp{\asppv})\}$ $=$ $\{\conc{E} \mid E \in \semv(\grounddlp{\asppv})\}$ for $\semv \in \semlistc$. 
\end{theorem}

\section{Evaluation}
To demonstrate feasibility of our grounding approach we conducted experiments on three distinct datasets~\cite{monterosso_2025_16875768}. 
To evaluate the feasibility of reasoning, we used in scenario S1 F.O. ASPIC+ instances inspired by~\cite{legalBench},
and in scenario S2 F.O. ASPIC+ instances with increasing number of atoms, rules and variables. 
In scenario S3 we assess the quality of the groundings by translating well known ASP instances~\cite{aspComp} to ASPIC+ instances. The experiments were conducted on 
an Intel(R) Xeon(R) Platinum $8470$, $3.5$-GHz, $1.0$-TB RAM.
For each instance we impose a $20$-minute time limit for grounding and in scenarios S1 and S2 additionally $20$ minutes for reasoning.   Additionally, we impose a memory limit of $32G$.

\noindent
\textbf{Solvers.}
The proposed approach (including the simplifications) has been implemented in the prototypical grounder \angry~\cite{monterosso_2025_16872941}, which is written in Rust and makes use of the
Datalog engine \nemo~\cite{ivliev2024nemo}. To evaluate the use of \angry as part of a ground+solve pipeline for F.O. ASPIC+ instances we combine \angry~with 
\aspforaspic~\cite{Aspforaspic1}, an ASP-based reasoner for propositional ASPIC+. For the backend of \aspforaspic~we make use of 
\clingo~v5.7.1.~\cite{clingo}.
As a reference system for scenario S1 we use \argtp~\cite{CalegariOPS22}, the only argumentation tool we are aware of that handles an 
F.O. ASPIC+-like syntax.
For scenario S3 we compare against the well-known grounder \gringo~\cite{gringo}.

\noindent
\textbf{Scenario S1.}
The primary goal is to evaluate the use of our grounding approach as part of a ground+solve strategy for reasoning over F.O. ASPIC+ instances. This involves first grounding
ASPIC+ instances and in a second step solving the obtained propositional argumentation theory. More precisely, we focus on the 
enumeration of claim sets
 under the complete semantics, as this is the reasoning task the two solvers \aspforaspic~and \argtp~are capable of. 
 The instances consist of  ASPIC+ encodings denoted by Legal bench (inspired by~\cite{legalBench}).
We generated 
$20$ instances with an increasing number of atoms in the knowledge base for each domain. 
We compare our grounder \angry~together with \aspforaspic~against \argtp. Specifically, we compare the execution time of each system
 to complete the enumeration of claim sets under complete semantics.
Additionally, we compare the claim sets obtained by the two system to verify the correctness of our grounding. 

\noindent
\textbf{Scenario S2.}
Here we to evaluate the scalability of \angry. To this end we randomly generated F.O. ASPIC+ instances with increasing 
number of atoms, rules and variables. 
In each generated instance, some parameters are fixed. Constants appearing in the instances are randomly selected integers
 in the range $[0, 300]$. The arity of the literals, both in rules and in the atom base, is randomly chosen between $1$ and $5$, 
 with a probability of $80\%$ to fall between $1$ and $3$. The number of literals appearing in each rule is randomly picked 
 from $1$ to $10$, with $80\%$ probability assigned to values between $1$ and $4$. The number of literals involved in 
 contrary relations is randomly selected between $1$ and $3$.

Based on this fixed set of parameters, we define four different configurations by varying the number of strict (str.) and defeasible (def.)
rules, as well as the number of contrary relations included in the encoding. The configurations are defined as follows:
\begin{itemize}[itemsep=0pt, topsep=0pt]
    \item[$c_1$:] $10$ rules ($5$ str.\ and $5$ def.) and $7$ contrary relations
    \item[$c_2$:] $20$ rules ($10$ str.\ and $10$ def.) and $15$ contrary relations
    \item[$c_3$:] $40$ rules ($20$ str.\ and $20$ def.) and $25$ contrary relations
    \item[$c_4$:] $80$ rules ($40$ str.\ and $40$ def.) and $30$ contrary relations
    \item[$c_5$:] $160$ rules ($80$ str.\ and $80$ def.) and $90$ contrary relations
\end{itemize}
For each configuration, we generate $7$ sets of instances, each with a maximum number of variable symbols per rule, 
incrementing from $3$ to $13$ for each set. For every configuration and  number of variable symbols, we 
generate $20$ instances, where the number of atoms in the atom base increases from $43200$ to $655584$.  This setup allows us to systematically analyze the scalability of \angry~under increasing structural complexity of the instances.

\noindent
\textbf{Scenario S3.} Here the aim is to study the extent to which our grounder meets the performance baseline of existing ASP grounders when used on ASP programs, translated to ASPIC+ theories. 
We compare the size of the grounding produced by \gringo~in the ASP domain, with those obtained by \angry~for the corresponding ASPIC+ instances, which are derived by translating the ASP instances used for \gringo~into ASPIC+.  The instances consist of three problem classes from the sixth ASP competition~\cite{aspComp}, 
namely StableMarriage, VisitAll and GraphColouring. In particular, we use all of the $20$ instances of each problem class. 
We apply the existing translations~\cite{abaLpEquivalence} and~\cite{Heyninck19} 
on the ASP encodings, obtaining the argument theories given as input to \angry. These translations
 were originally designed for propositional programs, therefore we slightly adapted them to handle variables.
Since the default output format for the two systems \gringo~and \angry~is  different, the groundings were generated using the \emph{text} option, which produces the groundings in a human-readable 
format and is very similar for both systems. Additionally, we measure and compare the grounding times of both systems. In this case, 
the output format used is the standard one.

\begin{table*}[ht]
    \centering
    \begin{small}
    \begin{tabular}{llccccccc}
        \toprule
        \multicolumn{2}{l}{Config} & 3v & 4v & 5v & 7v & 9v & 11v & 13v \\
        \midrule
        \multirow{2}{*}{$c_1$}
				            & grounded / solved        & 16 / 16 & 16 / 16 & 15 / 15 & 15 / 13 & 9 / 7   & 7 / 7   & 5 /4 \\
            & TO / MO                 & 4 / 0   & 4 / 0   & 5 / 0   & 3 / 2   & 7 / 4   & 7 / 6   & 5 / 10  \\
					  & mean time (\emph{s})    & 261.1   & 181.4   & 187.0   & 228.1   & 313.8   & 181.5   & 406.4   \\
            & median time (\emph{s})  & 117.5   & 81.8    & 71.3    & 141.9   & 129.7   & 139.6   & 278.4   \\

				\midrule
        \multirow{2}{*}{$c_2$} 
				            & grounded / solved        & 16 / 16 & 15 / 15 & 15 / 15 & 9 / 7   & 7 / 5   & 4 / 3   & 2 / 0   \\
            & TO / MO                 & 4 / 0   & 5 / 0   & 5 / 0   & 9 / 2   & 11 / 2  & 9 / 7   & 7 / 11  \\
            & mean time (\emph{s})    & 284.3   & 259.1   & 336.1   & 418.6   & 488.4   & 580.8   & 731.2   \\
            & median time (\emph{s})  & 107.4   & 79.4    & 147.5   & 313.6   & 632.2   & 665.8   & 731.2   \\

        \midrule
				\multirow{2}{*}{$c_3$} 
				    & grounded / solved        & 14 / 13 & 13 / 13 & 11 / 11 & 5 / 4   & 2 / 2   & 2 / 2  & 0 / 0  \\
            & TO / MO                 & 6 / 0   & 7 / 0   & 9 / 0   & 12 / 3  & 8 / 10  & 9 / 9 & 8 / 12 \\
            & mean time (\emph{s})    & 238.6   & 258.1   & 250.9   & 409.4   & 557.0   & 336.7 & – \\
            & median time (\emph{s})  & 144.7   & 237.9   & 231.7   & 353.1   & 557.0   & 336.7 & – \\


				\midrule
				\multirow{2}{*}{$c_4$} 
				    & grounded / solved        & 15 / 15 & 13 / 13 & 8 / 8   & 0 / 0   & –       & –  & –     \\
            & TO / MO                 & 5 / 0   & 7 / 0   & 9 / 3   & 6 / 14  & –       & –    & –   \\
            & mean time (\emph{s})    & 274.6   & 360.2   & 503.4   & –       & –       & –   & –    \\
            & median time (\emph{s})  & 54.3    & 237.6   & 215.3   & –       & –       & –   & –    \\


				\midrule
				\multirow{2}{*}{$c_5$} 
				    & grounded / solved        & 10 / 7  & 6 / 5   & 3 / 3   & 0 / 0   & –       & –   & –     \\
            & TO / MO                 & 6 / 4   & 9 / 5   & 6 / 11  & 9 / 11  & –       & –   & –     \\
            & mean time (\emph{s})    & 170.2   & 373.2   & 576.4   & –       & –       & –   & –     \\
            & median time (\emph{s})  & 127.3   & 331.7   & 486.9   & –       & –       & –   & –     \\


					\bottomrule
   \end{tabular}
    \end{small}
    \caption{Results of scenario S2. Configurations $c_1, \dots ,c_5$ are as described in S2; the columns 3v, 4v, $\dots$ denote up to 
    3 (4, $\dots$) variables per rule; TO stands for time out and MO for memory out, both refer to the grounding step. Mean time and 
    median time both refer to the execution time of the grounding step, reported in seconds.}
 \label{tab:random_bench_res}
\end{table*}

\noindent
\textbf{Expectations.}
\begin{enumerate}[itemsep=0pt, topsep=0pt]
\item[(E1)] \angry+\aspforaspic~will outperform \argtp~(which has struggled on larger instances in the evaluation in~\cite{legalBench});
 \item[(E2)] \angry+\aspforaspic~ and \argtp~output the same claim sets for the same instance;
 \item[(E3)] grounding time increases with the size of the grounding;
 \item[(E4)] \angry~can handle instances of realistic size;
 \item[(E5)] \angry~will produce slightly larger groundings compared to \gringo, due to additional rules needed for the translation from ASP to ASPIC+;
 \item[(E6)] \angry~will not outperform \gringo~on ASP instances as \gringo~is a very efficient system that introduces several 
 extra optimizations 
for the ASP domain. 
\end{enumerate}
\begin{figure}[ht]
        \centering
        \includegraphics[width=\linewidth]{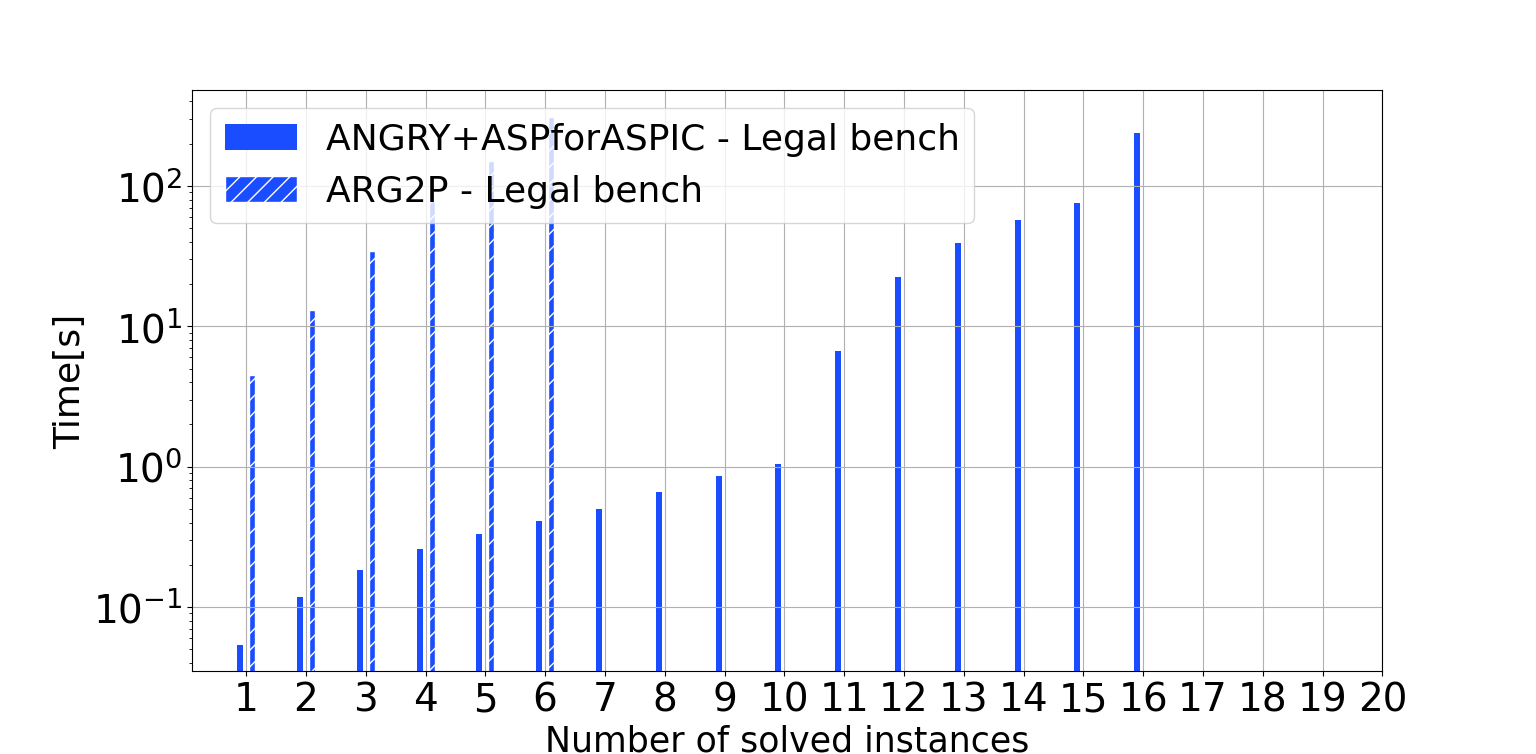}
        \caption{Results of scenario S1 comparing \angry+\aspforaspic~vs. \argtp.}
        \label{fig:angry_vs_arg2p}
        \end{figure}
       
\begin{figure}[ht]
      \centering
        \includegraphics[width=\linewidth]{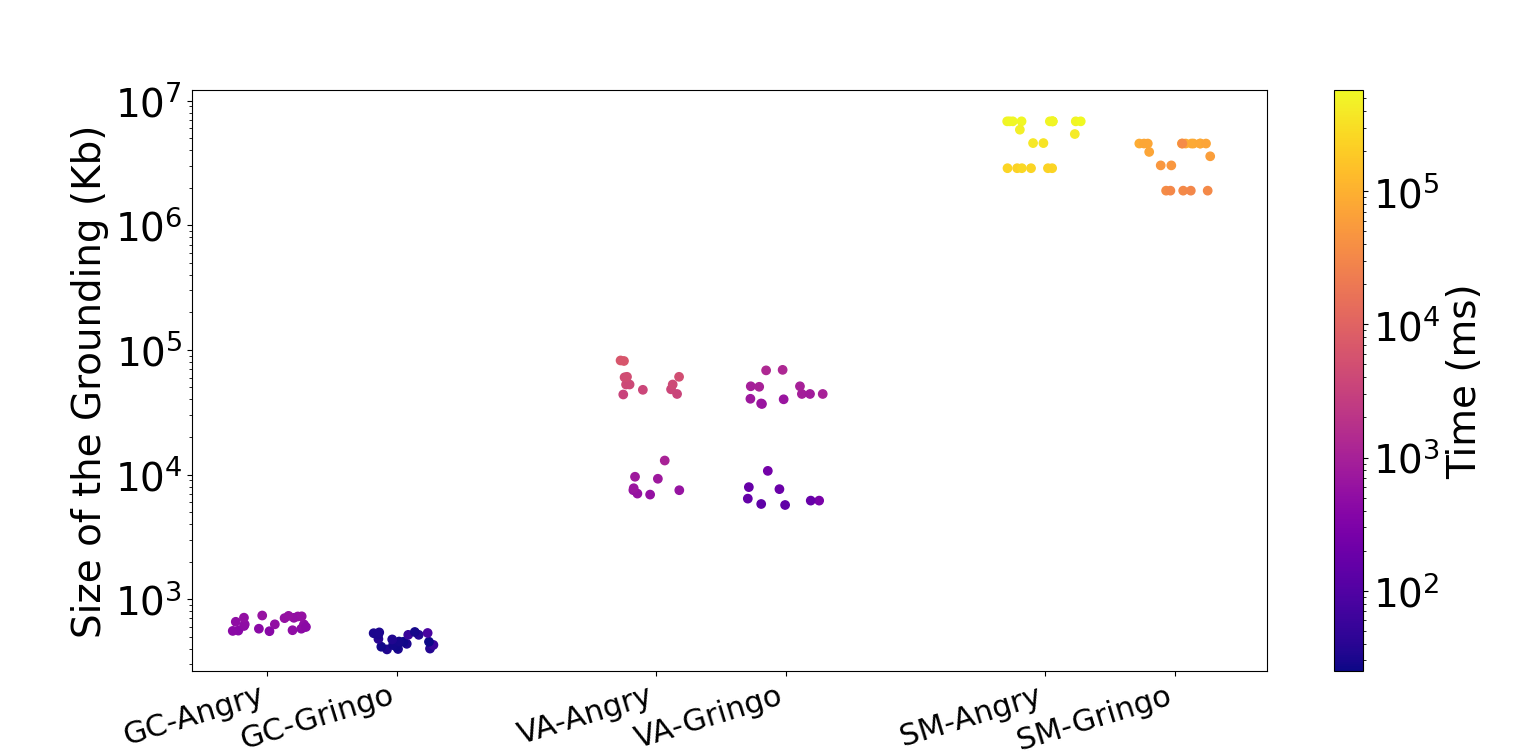}
        \caption{Results of scenario S3 comparing \angry~vs \gringo, where GC, VA and SM stand for GraphColouring,  VisitAll and StableMarriage.}
        \label{fig:angry_vs_gringo}
\end{figure}

\noindent
\textbf{Observations and Results.}
Figure~\ref{fig:angry_vs_arg2p} confirms (E1). We use a cactus plot, which shows the time that is at 
least required to solve a specific number of instances. Thus, the first bar shows the time for the fastest solving, the second bar 
the sum of the solving time for the two fastest instances, and so on. We report the time required by both the system \angry+\aspforaspic~and \argtp~to process the instances of scenario S1. From the plot, it is evident that \angry+\aspforaspic~outperforms
 \argtp~in terms of efficiency for these instances. In fact, our system solves $16$ instances from LEGAL BENCH, 
 whereas \argtp~only solves $6$ instances within the time limit. For these instances we were able to confirm (E2) by assessing equivalence of the results. 
 
Table~\ref{tab:random_bench_res} shows the results of scenario S2, where we report the number of grounded and solved instances 
in the given time, as well as the mean and median time for the grounding step.
 For those instances where the grounding step did not
finish, we report if they run into a time out  or ran out of memory. We can observe that in general the number of variables in the rules
has the largest influence on the grounding, which is not a big surprise. Increasing the number of rules and contraries, is not a big problem,
as long as the number of variables is low. For challenging instances we observe more instances failing due to memory outage than due to
time out. Concluding on scenario S2 we see (E4) confirmed as \angry~is able to ground instances of reasonable size and with a reasonable
number of variables in rules. We assume that most real world instances would not require to use more than 4 or 5 variables within one rule.
 
Finally, Figure~\ref{fig:angry_vs_gringo} confirms (E3), (E5) and (E6). There, we report the grounding size obtained by both \angry~and \gringo~for each instance from scenario S3. The grounding size obtained by \angry~is, as expected, comparable and 
slightly larger than that by \gringo. This outcome confirms the quality of \angry's grounding, as it does not contain 
unnecessary information. Additionally, the grounding time is reported via a color scale. As anticipated, our system is less efficient 
than \gringo. In this regard, further efforts will be made in the future to improve the system's efficiency.

\section{Conclusion}
We proposed a novel grounding approach for rule-based argumentation that makes use of a Datalog engine. 
Optimizations performed in the grounding avoid the construction of unnecessary parts and thus produces an intelligent grounding that
preserves acceptable claims.
Our experimental analysis demonstrates feasibility of our prototype grounder \angry, which makes use of the Datalog engine \nemo~\cite{ivliev2024nemo}. 
In particular, the use of \angry~together with the system \aspforaspic~\cite{Aspforaspic2} as part of ground+solve approach 
for reasoning over ASPIC+ instances clearly outperforms the only existing argumentation-based system that we are aware of 
that supports a first-order ASPIC+-like syntax.  On the other hand, the use of our grounder as an ASP grounder 
(via translations from ASP to ASPIC+~\cite{Heyninck19,abaLpEquivalence}) shows that our grounder meets the 
performance baseline of the ASP grounder \gringo~\cite{gringo}, although there is also some loss of performance 
due to the translations and given the additional optimizations of \gringo~for ASP. 

\noindent
\textbf{Future Work.} We plan to extend our approach to grounding to also handle preferences, which are a further crucial aspect of frameworks for rule-based argumentation such as ABA (see e.g.\cite{CyrasT16,Wakaki17} and ASPIC+~\cite{ModgilP18}.  On the other hand, we want to study to what extent additional constructs that are commonly used in ASP such as conditional literals, external atoms or aggregates~\cite{GebserS16} can meaningfully be incorporated. Additional efforts will also be devoted to improving the system's efficiency.  This includes, evaluating the use of different Datalog and ASP (used as Datalog) grounders as the back-end of our system as well as the combination of our grounder with different reasoning approaches also for ABA (e.g.~\cite{LehtonenWJ21,DiGaGo22}).

\section*{Acknowledgments}This work is partially supported by BMFTR (German Federal Ministry of Research, Technology and Space) in project \href{https://semeco.info}{Semeco} (Secure medical microsystems and communications) under grant 03ZU1210B. We also acknowledge support by BMFTR in DAAD project 57616814 (\href{https://secai.org/}{SECAI, School of Embedded Composite AI}) as part of the program Konrad Zuse Schools of Excellence in Artificial Intelligence. Moreover, we acknowledge partial financial support from MUR project PRIN 2022 EPICA (CUP H53D23003660006) funded by the European Union - Next Generation EU and from the PNRR project FAIR - Future AI Research (PE00000013), Spoke 9 -Green-aware AI, under the NRRP MUR program funded by the NextGenerationEU.  Finally, we acknowledge the computing time made available to us on the high-performance computer at the NHR Center of TU Dresden. This center is jointly supported by the BMFTR and the German state governments participating in the \href{www.nhr-verein.de/unsere-partner}{NHR}.
\bibliographystyle{kr}
\bibliography{mybib}

\begin{thebibliography}{}

\bibitem[\protect\citeauthoryear{Abiteboul, Hull, and
  Vianu}{1995}]{abiteboul1995}
Abiteboul, S.; Hull, R.; and Vianu, V.
\newblock 1995.
\newblock {\em Foundations of Databases}.
\newblock Reading, Massachusetts: Addison-Wesley.

\bibitem[\protect\citeauthoryear{Besin, Hecher, and Woltran}{2023}]{BesinHW23}
Besin, V.; Hecher, M.; and Woltran, S.
\newblock 2023.
\newblock On the structural complexity of grounding - tackling the {ASP}
  grounding bottleneck via epistemic programs and treewidth.
\newblock In {\em {ECAI}}, volume 372 of {\em Frontiers in Artificial
  Intelligence and Applications},  247--254.
\newblock {IOS} Press.

\bibitem[\protect\citeauthoryear{Besnard \bgroup et al\mbox.\egroup
  }{2014}]{BesnardGHMPST14}
Besnard, P.; Garc{\'{\i}}a, A.~J.; Hunter, A.; Modgil, S.; Prakken, H.; Simari,
  G.~R.; and Toni, F.
\newblock 2014.
\newblock Introduction to structured argumentation.
\newblock {\em Argument Comput.} 5(1):1--4.

\bibitem[\protect\citeauthoryear{Bondarenko \bgroup et al\mbox.\egroup
  }{1997}]{BondarenkoDKT97}
Bondarenko, A.; Dung, P.~M.; Kowalski, R.~A.; and Toni, F.
\newblock 1997.
\newblock An abstract, argumentation-theoretic approach to default reasoning.
\newblock {\em Artif. Intell.} 93:63--101.

\bibitem[\protect\citeauthoryear{Bondarenko, Toni, and
  Kowalski}{1993}]{BondarenkoTK93}
Bondarenko, A.; Toni, F.; and Kowalski, R.~A.
\newblock 1993.
\newblock An assumption-based framework for non-monotonic reasoning.
\newblock In Pereira, L.~M., and Nerode, A., eds., {\em Proceedings of the 2nd
  International Workshop on Logic Programming and Non-monotonic Reasoning
  (LPNMR 1993)},  171--189.
\newblock {MIT} Press.

\bibitem[\protect\citeauthoryear{Calegari \bgroup et al\mbox.\egroup
  }{2022}]{CalegariOPS22}
Calegari, R.; Omicini, A.; Pisano, G.; and Sartor, G.
\newblock 2022.
\newblock {Arg2P}: an argumentation framework for explainable intelligent
  systems.
\newblock {\em J. Log. Comput.} 32(2):369--401.

\bibitem[\protect\citeauthoryear{Calimeri \bgroup et al\mbox.\egroup
  }{2017}]{calimeri2017dlv}
Calimeri, F.; Fusc{\`a}, D.; Perri, S.; and Zangari, J.
\newblock 2017.
\newblock {I-DLV}: the new intelligent grounder of {DLV}.
\newblock {\em Intelligenza Artificiale} 11(1):5--20.

\bibitem[\protect\citeauthoryear{Caminada and Schulz}{2017}]{abaLpEquivalence}
Caminada, M., and Schulz, C.
\newblock 2017.
\newblock On the equivalence between assumption-based argumentation and logic
  programming.
\newblock {\em Journal of Artificial Intelligence Research} 60:779--825.

\bibitem[\protect\citeauthoryear{Ciatto, Calegari, and
  Omicini}{2021}]{CiattoCO21}
Ciatto, G.; Calegari, R.; and Omicini, A.
\newblock 2021.
\newblock {2P-Kt}: {A} logic-based ecosystem for symbolic {AI}.
\newblock {\em SoftwareX} 16:100817.

\bibitem[\protect\citeauthoryear{Craven and Toni}{2016}]{CravenT16}
Craven, R., and Toni, F.
\newblock 2016.
\newblock Argument graphs and assumption-based argumentation.
\newblock {\em Artif. Intell.} 233:1--59.

\bibitem[\protect\citeauthoryear{Cyras and Toni}{2016}]{CyrasT16}
Cyras, K., and Toni, F.
\newblock 2016.
\newblock {ABA+:} assumption-based argumentation with preferences.
\newblock In {\em {KR}},  553--556.

\bibitem[\protect\citeauthoryear{Diller, Gaggl, and
  Gorczyca}{2021}]{DillerGG21}
Diller, M.; Gaggl, S.~A.; and Gorczyca, P.
\newblock 2021.
\newblock Flexible dispute derivations with forward and backward arguments for
  assumption-based argumentation.
\newblock In {\em {CLAR}}, volume 13040 of {\em LNCS},  147--168.
\newblock Springer.

\bibitem[\protect\citeauthoryear{Diller, Gaggl, and Gorczyca}{2022}]{DiGaGo22}
Diller, M.; Gaggl, S.~A.; and Gorczyca, P.
\newblock 2022.
\newblock Strategies in flexible dispute derivations for assumption-based
  argumentation.
\newblock In {\em SAFA@COMMA}, volume 3236 of {\em {CEUR} Workshop
  Proceedings},  59--72.

\bibitem[\protect\citeauthoryear{Dung and Thang}{2014}]{DungT14}
Dung, P.~M., and Thang, P.~M.
\newblock 2014.
\newblock Closure and consistency in logic-associated argumentation.
\newblock {\em J. Artif. Intell. Res.} 49:79--109.

\bibitem[\protect\citeauthoryear{Dung}{1995}]{Dung95}
Dung, P.~M.
\newblock 1995.
\newblock On the acceptability of arguments and its fundamental role in
  nonmonotonic reasoning, logic programming and n-person games.
\newblock {\em Artif. Intell.} 77(2):321--358.

\bibitem[\protect\citeauthoryear{East and Truszczynski}{2000}]{EastT00}
East, D., and Truszczynski, M.
\newblock 2000.
\newblock {DATALOG} with constraints - an answer-set programming system.
\newblock In {\em {AAAI/IAAI}},  163--168.
\newblock {AAAI} Press / The {MIT} Press.

\bibitem[\protect\citeauthoryear{Faber}{2020}]{Faber20}
Faber, W.
\newblock 2020.
\newblock An introduction to answer set programming and some of its extensions.
\newblock In {\em {RW}}, volume 12258 of {\em Lecture Notes in Computer
  Science},  149--185.
\newblock Springer.

\bibitem[\protect\citeauthoryear{Garc{\'{\i}}a and Simari}{2004}]{GarciaS04}
Garc{\'{\i}}a, A.~J., and Simari, G.~R.
\newblock 2004.
\newblock Defeasible logic programming: An argumentative approach.
\newblock {\em Theory Pract. Log. Program.} 4(1-2):95--138.

\bibitem[\protect\citeauthoryear{Garc{\'{\i}}a, Prakken, and
  Simari}{2020}]{GarciaPS20}
Garc{\'{\i}}a, A.~J.; Prakken, H.; and Simari, G.~R.
\newblock 2020.
\newblock A comparative study of some central notions of {ASPIC+} and {DeLP}.
\newblock {\em Theory Pract. Log. Program.} 20(3):358--390.

\bibitem[\protect\citeauthoryear{Gebser and Schaub}{2016}]{GebserS16}
Gebser, M., and Schaub, T.
\newblock 2016.
\newblock Modeling and language extensions.
\newblock {\em {AI} Mag.} 37(3):33--44.

\bibitem[\protect\citeauthoryear{Gebser \bgroup et al\mbox.\egroup
  }{2015}]{gringo}
Gebser, M.; Harrison, A.; Kaminski, R.; Lifschitz, V.; and Schaub, T.
\newblock 2015.
\newblock Abstract gringo.
\newblock {\em Theory and Practice of Logic Programming} 15(4-5):449--463.

\bibitem[\protect\citeauthoryear{Gebser \bgroup et al\mbox.\egroup
  }{2016}]{clingo}
Gebser, M.; Kaminski, R.; Kaufmann, B.; Ostrowski, M.; Schaub, T.; and Wanko,
  P.
\newblock 2016.
\newblock Theory solving made easy with clingo 5.
\newblock In {\em Technical Communications of the 32nd International Conference
  on Logic Programming (ICLP 2016)}.
\newblock Schloss-Dagstuhl-Leibniz Zentrum f{\"u}r Informatik.

\bibitem[\protect\citeauthoryear{Gebser, Maratea, and Ricca}{2015}]{aspComp}
Gebser, M.; Maratea, M.; and Ricca, F.
\newblock 2015.
\newblock The design of the sixth answer set programming competition.
\newblock In Calimeri, F.; Ianni, G.; and Truszczynski, M., eds., {\em Logic
  Programming and Nonmonotonic Reasoning},  531--544.
\newblock Cham: Springer International Publishing.

\bibitem[\protect\citeauthoryear{Heyninck}{2019}]{Heyninck19}
Heyninck, J.
\newblock 2019.
\newblock Relations between assumption-based approaches in nonmonotonic logics
  and formal argumentation.
\newblock {\em {FLAP}} 6(2):319--360.

\bibitem[\protect\citeauthoryear{Ivliev \bgroup et al\mbox.\egroup
  }{2024}]{ivliev2024nemo}
Ivliev, A.; Gerlach, L.; Meusel, S.; Steinberg, J.; and Kr{\"{o}}tzsch, M.
\newblock 2024.
\newblock Nemo: Your friendly and versatile rule reasoning toolkit.
\newblock In Marquis, P.; Ortiz, M.; and Pagnucco, M., eds., {\em Proceedings
  of the 21st International Conference on Principles of Knowledge
  Representation and Reasoning (KR 2024)},  743--754.
\newblock IJCAI Organization.

\bibitem[\protect\citeauthoryear{Jordan, Scholz, and
  Subotic}{2016}]{JordanSS16}
Jordan, H.; Scholz, B.; and Subotic, P.
\newblock 2016.
\newblock Souffl{\'{e}}: On synthesis of program analyzers.
\newblock In {\em {CAV} {(2)}}, volume 9780 of {\em Lecture Notes in Computer
  Science},  422--430.
\newblock Springer.

\bibitem[\protect\citeauthoryear{Järvisalo, Lehtonen, and
  Niskanen}{2025}]{jarvisaloln25}
Järvisalo, M.; Lehtonen, T.; and Niskanen, A.
\newblock 2025.
\newblock {ICCMA} 2023: 5th international competition on computational models
  of argumentation.
\newblock {\em Artificial Intelligence} 342:104--311.

\bibitem[\protect\citeauthoryear{Lehtonen \bgroup et al\mbox.\egroup
  }{2023}]{LehtonenR0W23}
Lehtonen, T.; Rapberger, A.; Ulbricht, M.; and Wallner, J.~P.
\newblock 2023.
\newblock Argumentation frameworks induced by assumption-based argumentation:
  Relating size and complexity.
\newblock In {\em {KR}},  440--450.

\bibitem[\protect\citeauthoryear{Lehtonen \bgroup et al\mbox.\egroup
  }{2024a}]{LehtonenOWJ24}
Lehtonen, T.; Odekerken, D.; Wallner, J.~P.; and J{\"{a}}rvisalo, M.
\newblock 2024a.
\newblock Complexity results and algorithms for preferential argumentative
  reasoning in {ASPIC+}.
\newblock In {\em {KR}}.

\bibitem[\protect\citeauthoryear{Lehtonen \bgroup et al\mbox.\egroup
  }{2024b}]{LehtonenRT0W24}
Lehtonen, T.; Rapberger, A.; Toni, F.; Ulbricht, M.; and Wallner, J.~P.
\newblock 2024b.
\newblock Instantiations and computational aspects of non-flat assumption-based
  argumentation.
\newblock In {\em {IJCAI}},  3457--3465.
\newblock ijcai.org.

\bibitem[\protect\citeauthoryear{Lehtonen, Wallner, and
  J{\"{a}}rvisalo}{2021}]{LehtonenWJ21}
Lehtonen, T.; Wallner, J.~P.; and J{\"{a}}rvisalo, M.
\newblock 2021.
\newblock Declarative algorithms and complexity results for assumption-based
  argumentation.
\newblock {\em J. Artif. Intell. Res.} 71:265--318.

\bibitem[\protect\citeauthoryear{Lehtonen, Wallner, and
  J{\"a}rvisalo}{2022}]{Aspforaspic2}
Lehtonen, T.; Wallner, J.; and J{\"a}rvisalo, M.
\newblock 2022.
\newblock {ASPforASPIC}: {ASP}-based algorithms for abstract rule-based
  argumentation (aspic+).

\bibitem[\protect\citeauthoryear{Lehtonen, Wallner, and
  Järvisalo}{2020}]{Aspforaspic1}
Lehtonen, T.; Wallner, J.~P.; and Järvisalo, M.
\newblock 2020.
\newblock {An Answer Set Programming Approach to Argumentative Reasoning in the
  ASPIC+ Framework}.
\newblock In {\em {Proceedings of the 17th International Conference on
  Principles of Knowledge Representation and Reasoning}},  636--646.

\bibitem[\protect\citeauthoryear{Marek and Truszczynski}{1999}]{MarekT99}
Marek, V.~W., and Truszczynski, M.
\newblock 1999.
\newblock Stable models and an alternative logic programming paradigm.
\newblock In {\em The Logic Programming Paradigm}, Artificial Intelligence.
  Springer.
\newblock  375--398.

\bibitem[\protect\citeauthoryear{Modgil and Prakken}{2018}]{ModgilP18}
Modgil, S., and Prakken, H.
\newblock 2018.
\newblock Abstract rule-based argumentation.
\newblock In Baroni, P.; Gabbay, D.; and Giacomin, M., eds., {\em Handbook of
  Formal Argumentation}. College Publications.
\newblock  287--364.

\bibitem[\protect\citeauthoryear{Monterosso \bgroup et al\mbox.\egroup
  }{2025a}]{monterosso_2025_16872941}
Monterosso, G.; Gaggl, S.~A.; Diller, M.; Hanisch, P.; and Rauschenbach, F.
\newblock 2025a.
\newblock {ANGRY}: A grounder for rule-based argumentation.
\newblock https://doi.org/10.5281/zenodo.16872941.

\bibitem[\protect\citeauthoryear{Monterosso \bgroup et al\mbox.\egroup
  }{2025b}]{monterosso_2025_16875768}
Monterosso, G.; Gaggl, S.~A.; Diller, M.; Hanisch, P.; and Rauschenbach, F.
\newblock 2025b.
\newblock Benchmark instances for {ANGRY}: A grounder for rule-based
  argumentation.
\newblock https://doi.org/10.5281/zenodo.16875768.

\bibitem[\protect\citeauthoryear{Nenov \bgroup et al\mbox.\egroup
  }{2015}]{NenovPMHWB15}
Nenov, Y.; Piro, R.; Motik, B.; Horrocks, I.; Wu, Z.; and Banerjee, J.
\newblock 2015.
\newblock {RDFox}: {A} highly-scalable {RDF} store.
\newblock In {\em {ISWC} {(2)}}, volume 9367 of {\em Lecture Notes in Computer
  Science},  3--20.
\newblock Springer.

\bibitem[\protect\citeauthoryear{Niemel{\"{a}}}{1999}]{Niemela99}
Niemel{\"{a}}, I.
\newblock 1999.
\newblock Logic programs with stable model semantics as a constraint
  programming paradigm.
\newblock {\em Ann. Math. Artif. Intell.} 25(3-4):241--273.

\bibitem[\protect\citeauthoryear{Odekerken \bgroup et al\mbox.\egroup
  }{2023}]{OdekerkenLBWJ23}
Odekerken, D.; Lehtonen, T.; Borg, A.; Wallner, J.~P.; and J{\"{a}}rvisalo, M.
\newblock 2023.
\newblock Argumentative reasoning in {ASPIC+} under incomplete information.
\newblock In {\em {KR}},  531--541.

\bibitem[\protect\citeauthoryear{Popescu and Wallner}{2023}]{PopescuW23}
Popescu, A., and Wallner, J.~P.
\newblock 2023.
\newblock Reasoning in assumption-based argumentation using
  tree-decompositions.
\newblock In {\em {JELIA}}, volume 14281 of {\em Lecture Notes in Computer
  Science},  192--208.
\newblock Springer.

\bibitem[\protect\citeauthoryear{Prakken}{2010}]{Prakken10}
Prakken, H.
\newblock 2010.
\newblock An abstract framework for argumentation with structured arguments.
\newblock {\em Argument Comput.} 1(2):93--124.

\bibitem[\protect\citeauthoryear{Robaldo \bgroup et al\mbox.\egroup
  }{2024}]{legalBench}
Robaldo, L.; Batsakis, S.; Calegari, R.; Calimeri, F.; Fujita, M.; Governatori,
  G.; Morelli, M.~C.; Pacenza, F.; Pisano, G.; Satoh, K.; et~al.
\newblock 2024.
\newblock Compliance checking on first-order knowledge with conflicting and
  compensatory norms: a comparison among currently available technologies.
\newblock {\em Artificial Intelligence and Law} 32(2):505--555.

\bibitem[\protect\citeauthoryear{Urbani, Jacobs, and
  Kr{\"{o}}tzsch}{2016}]{UrbaniJK16}
Urbani, J.; Jacobs, C. J.~H.; and Kr{\"{o}}tzsch, M.
\newblock 2016.
\newblock Column-oriented datalog materialization for large knowledge graphs.
\newblock In {\em {AAAI}},  258--264.
\newblock {AAAI} Press.

\bibitem[\protect\citeauthoryear{Wakaki}{2017}]{Wakaki17}
Wakaki, T.
\newblock 2017.
\newblock Assumption-based argumentation equipped with preferences and its
  application to decision making, practical reasoning, and epistemic reasoning.
\newblock {\em Comput. Intell.} 33(4):706--736.

\bibitem[\protect\citeauthoryear{Weinzierl, Taupe, and
  Friedrich}{2020}]{WeinzierlTF20}
Weinzierl, A.; Taupe, R.; and Friedrich, G.
\newblock 2020.
\newblock Advancing lazy-grounding {ASP} solving techniques - restarts, phase
  saving, heuristics, and more.
\newblock {\em Theory Pract. Log. Program.} 20(5):609--624.

\end{thebibliography}

\newpage
\appendix
\section{Appendix}\label{sec:05}

\subsection{Preliminaries}

Towards proving the main theorems of our work we start by introducing some further notions and notation.  First, apart from the 
notions about arguments as defined in Section~\ref{subsec-prop-aspic}. 
we need the concepts of the sub-arguments and weak points  of an argument:

\begin{definition}\label{def:argsex}  Let $\asppv = \aspp$ be an \argpr~and $\argv = \stmv \in \kbsa \cup \kbso$ as well as $\argv' = \aarg{\argv_1,\ldots,\argv_m}{\stmv'}$ arguments of $\asppv$. Then, the \emph{subarguments} of $\argv$ are  $\subo{\argv} = \{\argv\}$, while $\subo{\argv'} = \{\argv'\} \cup \bigcup_{1\leq i \leq m} \subo{\argv_i}$.  Moreover, for any argument $\argv$ of $\asppv$, let the \emph{weak points} of the argument be $\wpa{\argv} = (\pre{\argv} \cap \kbso) \cup \bigcup_{\rulv \in \ruld} \defe{\rulv}$\footnote{This is only a slight variation of the notion $\defe{\argv}$ used in the main text, pinpointing the exact points at which an argument can be attacked.}.
\end{definition}

To be able to connect $\groundp{\asppv}$ and $\grounddlp{\asppv}$ (the latter for each of the different versions of the Datalog 
program $\datpva$ corresponding to a theory $\asppv$ we define) we introduce the notion of a \emph{derivation} for Datalog programs.  
This is essentially the notion of argument from  Section~\ref{subsec-prop-aspic} 
translated to the Datalog world:

\begin{definition} The set of \emph{derivations} of a ground Datalog program $\datpv$ is defined inductively as follows: {\bf i }) if $\grul{}{\hedv} \in \datpv$, then $\derv = \grul{}{\hedv}$ is derivation with \emph{conclusion} $\concd{\derv} = \hedv$; {\bf ii}) if $\derv_1,\ldots,\derv_m$ are derivations and $\grul{\conc{\derv_1},\ldots,\conc{\derv_m}}{\hedv} \in \datpv$, then $\derv = \grul{\derv_1,\ldots,\derv_m}{\hedv}$ is a derivation with $\conc{\derv} = \hedv$.  There are no other derivations than those defined by i) and ii).  
\end{definition}

\noindent We denote the set of derivations of a ground Datalog program $\datpv$ as $\ders{\datpv}$. It is easy to see then that for a Datalog program $\datpv$, $\qrv(\tlv) \in \gqr{\datpv}{\qrv}$ iff there is a derivation $\derv \in \ders{\groundp{\datpv}}$ with $\conc{\derv} = \qrv(\tlv)$.  

In the following proofs we will now often represent an argument $\argv$ for a (ground) argumentation theory $\asppv = \aspp$ as $\argd{b_1,\ldots,b_m,\rulv_1,\ldots,\rulv_n}$ ($m \geq 0$, $n \geq 0$, $m+n \geq 1$) where $\pre{\argv} = \{b_1,\ldots,b_m\}$ and $\rulsa{\argv} = \{\rulv_1,\ldots,\rulv_n\}$.  Analogously, we represent a derivation $\derv$ for a Datalog program $\datpv$ as $\datd{\rulv_1,\ldots,\rulv_n}$ ($n \geq 1$) where the $\rulv_i$ ($1 \leq i \leq n$) are exactly the rules used in the derivation $\derv$.  Thus, in this representation, we focus on the rules (and premisses in the case of arguments) used in the arguments and derivations, abstracting away from the internal structure (e.g. the order of application of the rules).    

To further ease our proofs, for each of the transformations of an argumentation theory $\asppv = \aspp$ into a Datalog program $\datpva$ presented in our work (transformations~\ref{def:transf1} and~\ref{def:transf2}), we introduce the notation $\hkvarp{1} = \hat{\kvar}$ for $\kvar \in \kbs$, while $\hrulvp{i}$ for $1 \leq i \leq 3$ and $\rulv \in \rul$ selects the first, second, or third respectively of the rules in $\hat{\rulv}$.  For instance, $\hrulvp{2} = \grul{\aux{\xlv}}{\ghedv{\ylv}}$ for a $\rulv = \srul{\abodv{\xlv}}{\ahedv{\ylv}} \in \ruls$ or $\rulv = \drul{\anamv{\zlv}}{\abodv{\xlv}}{\ahedv{\ylv}} \in \ruld$ for both Transformation~\ref{def:transf1} as well as Transformation~\ref{def:transf2}.  

We can now connect an argumentation theory $\asppv$ and its corresponding Datalog program $\datpva$ (via any of the transformations presented in our work) via their arguments and derivations.  Sepcifically, we have for an $\argv = \argd{\kvar_1,\ldots,\kvar_m,\rulv_{1}\sigma_{1},\ldots,\rulv_{l}\sigma_{l}} \in \arts{\agroundp{\asppv}}$ ($\kvar_i \in \kbs$, $\rulv_i \in \rul$, $\sigma_i: \vars \mapsto \heru{\asppv}$, $m\geq0$,$l\geq0$, $m+l\geq1$), let the corresponding proposed derivation $\derg{\argv}{i}$ for  $\groundp{\datpva}$ with $i \in \{1,2,3\}$ be $\derg{\argv}{i} = \datd{d_1,d_2,d_3}$ with $d_1 = \hkvarpt{1}{1},\ldots,\hkvarpt{m}{1}$, $d_2 =  \hrulvpt{1}{1}\sigma_{1},\hrulvpt{1}{2}\sigma_{1},\ldots,\hrulvpt{l-1}{1}\sigma_{l-1},\hrulvpt{l-1}{2}\sigma_{l-1}$, $d_3 = \hrulvpt{l}{1}\sigma_{l},\hrulvpt{l}{i}\sigma_{l}$ (note the $i$ in the definition of $d_3$).  On the other hand, for a derivation $\derg{\argv'}{i} \in \ders{\groundp{\datpva}}$ of the form $\derg{\argv'}{i} = \datd{d_1,d_2,d_3}$ with $d_1,d_2,d_3$ as above let the corresponding proposed argument for $\agroundp{\asppv}$  be $\argv' = \argd{\kvar_1,\ldots,\kvar_m,\rulv_{1}\sigma_{1},\ldots,\rulv_{l}\sigma_{l}}$. 

The following Lemma now holds independently of the transformation we consider in this work:
    
\begin{lemma}\label{lem:auxgrd1} Let $\asppv = \aspp$ be an \argth~and $\datpva$ the Datalog program for $\asppv$ as per Transformation~\ref{def:transf1} or Transformation~\ref{def:transf2}.  Then, $\kvar \in \agroundp{\asppv}$ iff $\hkvarp{1} \in \groundp{\datpva}$ for $\kvar \in \kbsa \cup \kbso$. Moreover, $\rulv\sigma \in \agroundp{\asppv}$ iff $\hrulvp{i}\sigma \in \groundp{\datpva}$ for $\rulv \in \rul$, $1 \leq i \leq 3$, and $\sigma$ a ground substitution $\vars \mapsto \heru{\asppv}$.
\end{lemma}

\begin{proof} For $\kvar \in \kbsa \cup \kbso$, $\kvar \in \asppv$ iff $\hatp{\kvar}{1} \in \datpva$ by construction. Hence, since $\kvar$ and $\hatp{\kvar}{1}$ are already ground, also $\kvar \in \agroundp{\asppv}$ iff $\hatp{\kvar}{1} \in \groundp{\datpva}$. For $\rulv \in \rul$, note, first of all, that $\asppv$ and $\datpva$ have the same Herbrand universe (\ie~$\aheru{\asppv} = \heru{\datpva}$).  Moreover, $\rulv \in \rul$ and $\hrulvp{i} \in \datpva$ for $1 \leq i \leq 3$ have the same variables.  Putting these two facts together one gets that $\rulv\sigma \in \agroundp{\asppv}$ iff $\hrulvp{i}\sigma \in \groundp{\datpva}$ for a ground substitution $\sigma$ and $1 \leq i \leq 3$.\end{proof}

\subsection{Proofs for Section~\ref{subsec:grd1}}\label{ap:prfsgrd1}


We now move to proving Lemma~\ref{lem:grd1} and Theorem~\ref{thm:grd1} from Section~\ref{subsec:grd1}.  For this, we first introduce the auxiliary lemmas~\ref{lem:auxgrd2} and~\ref{lem:auxgrd3}.

\begin{lemma}\label{lem:auxgrd2} Let $\asppv = \aspp$ be an \argth~and $\datpva$ de Datalog program for $\asppv$ as per Transformation~\ref{def:transf1}.   Then, {\bf i)}     $\argv \in \arts{\agroundp{\asppv}}$ iff 
$\derg{\argv}{1} \in \ders{\groundp{\datpva}}$; {\bf ii)}   $\argv \in \arts{\agroundp{\asppv}}$ with $\conc{\argv} = \stmv$ iff 
$\derg{\argv}{2} \in \ders{\groundp{\datpva}}$ with $\concd{\derg{\argv}{2}} = \stmv$; {\bf iii)} $\argv \in \arts{\agroundp{\asppv}}$ 
with $\pfua{\trl{\argv}} = \namv$ iff $\derg{\argv}{3} \in \ders{\groundp{\datpva}}$ with $\concd{\derg{\argv}{3}} = \namv$.     
\end{lemma}

\begin{proof} This follows from Lemma~\ref{lem:auxgrd1} and the structure of the arguments and derivations in the statement of the lemma.\end{proof}

\begin{lemma} \label{lem:auxgrd3} Let $\asppv = \aspp$ be an \argth, $\rulv \in \rul$, $\sigma: \vars \mapsto \heru{\asppv}$, $s \in \preds$ occurring in $\asppv$, and $\datpva$ the Datalog program for $\asppv$ as per Transformation~\ref{def:transf1}.   
Then,  i) $\auxt{\rulv}{\xlv}\sigma \in \gqr{\datpva}{\auxnr}$ iff there is a $\derg{\argv}{1} \in \ders{\groundp{\datpva}}$ for an 
$\argv \in \arts{\agroundp{\asppv}}$ with $\trl{\argv} = \rulv\sigma$; ii) $s({\xlv})\sigma \in \gqr{\datpva}{s}$ iff there is a 
$\derg{\argv}{2} \in \ders{\groundp{\datpva}}$ or $\derg{\argv}{3} \in \ders{\groundp{\datpva}}$  for an $\argv \in \arts{\agroundp{\asppv}}$ 
with $\conc{\argv} = s({\xlv})\sigma$ or $\pfua{\trl{\argv}} = s({\xlv})\sigma$.  
\end{lemma}

\begin{proof} This follows from the definitions of the queries ($\gqr{\datpva}{\auxnr}$ and $\gqr{\datpva}{s}$), the construction of $\datpva$, and Lemma~\ref{lem:auxgrd2}.\end{proof}

\begingroup
\renewcommand{\thelemma}{1} 
\begin{lemma}
    Let $\asppv$ be an \argth~and $\grounddlp{\asppv}$ the grounding via the Datalog program $\datpva$ as per Transformation~\ref{def:transf1}. Then, $\arts{\agroundp{\asppv}} = \arts{\grounddlp{\asppv}}$ and 
    $\atts{\agroundp{\asppv}} = \atts{\grounddlp{\asppv}}$.
    \end{lemma}
\endgroup
    
\begin{proof} 
    Let $\asppv = \aspp$.

    ($\arts{\agroundp{\asppv}} = \arts{\grounddlp{\asppv}}$)  We have that $\rulv\sigma \in \grounddlp{\rul}$ for a $\rulv \in \rul$ (and ground substitution $\sigma$) iff $\auxt{\rulv}{\xlv}\sigma \in \gqr{\datpva}{\auxnr}$ 
iff (by Lemma~\ref{lem:auxgrd3}) there is a $\derg{\argv}{1} \in \ders{\groundp{\datpva}}$ for an $\argv \in \arts{\agroundp{\asppv}}$
 with $\trl{\argv} = \rulv\sigma$. Since each $\rulv\sigma \in \rulsa{\arts{\agroundp{\asppv}}}$ is the top-rule of some 
 $\argv \in \arts{\agroundp{\asppv}}$ (the set of arguments is closed under sub-arguments), this means  $\rulv\sigma \in \grounddlp{\rul}$ 
 iff $\rulv\sigma \in \rulsa{\arts{\agroundp{\asppv}}}$.  But then, since also $\kbso \cup \kbsa \subseteq \grounddlp{\asppv}$ 
 and $\kbso \cup \kbsa \subseteq \agroundp{\asppv}$, exactly the same arguments can be constructed from $\agroundp{\asppv}$ 
 and $\grounddlp{\asppv}$, \ie~$\arts{\agroundp{\asppv}} = \arts{\grounddlp{\asppv}}$.

($\atts{\agroundp{\asppv}} \subseteq \atts{\grounddlp{\asppv}}$) Now consider an $\argv \in \arts{\agroundp{\asppv}}$ that attacks an $\argv' \in \arts{\agroundp{\asppv}}$. 
This means $\conc{\argv} \in S({\xlv})\sigma$ for a $s({\xlv})\sigma \in \wpa{\argv'}$ and $c: \acrul{\astmv{\xlv}}{\astmsv{\xlv}} \in \con$.
 But this means also there is a $\argv'' \in \subo{\argv'}$ with $\conc{\argv''} = s({\xlv})\sigma$ or $\pfua{\trl{\argv''}} = s({\xlv})\sigma$.  
 Then, there are $\derg{\argv''}{2} \in \ders{\groundp{\datpva}}$ or $\derg{\argv''}{3} \in \ders{\groundp{\datpva}}$ from where it follows, 
 by Lemma~\ref{lem:auxgrd3}, that $s({\xlv})\sigma \in \gqr{\datpva}{s}$ and, hence, $c\sigma \in \grounddlp{\con}$.  Since we already 
 know that each of the arguments $\argv$, $\argv'$, and $\argv''$ are also in $\arts{\grounddlp{\asppv}}$ we thus get that the attack 
 from $\argv$ to $\argv'$ is also in $\atts{\grounddlp{\asppv}}$. 

($\atts{\grounddlp{\asppv}} \subseteq \atts{\agroundp{\asppv}}$) That attacks between arguments $\argv,\argv'$ in $\grounddlp{\asppv}$ are preserved in $\agroundp{\asppv}$ follows, on the 
other hand, from the fact that such attacks are obtained via a $c\sigma \in \grounddlp{\con}$ for $c \in \con$ and 
$\sigma: \vars \mapsto \heru{\datpva}$. But, since $\heru{\datpva} = \heru{\asppv}$, then also $c\sigma \in \agroundp{\con}$ 
and, thus, $\argv$ also attacks $\argv'$ in $\agroundp{\asppv}$.\end{proof}

\begingroup
\renewcommand{\thetheorem}{1} 
\begin{theorem} 
Let $\asppv$ be an \argth~and $\grounddlp{\asppv}$ the grounding via the Datalog program $\datpva$ as per Transformation~\ref{def:transf1}. Then, $\semv(\asptv)$ $=$ $\semv(\agroundp{\asppv})$ $=$ $\semv(\grounddlp{\asppv})$ for $\semv \in \semlist$.
\end{theorem}
\endgroup

\begin{proof} For each $\semv \in \semlist$, $\semv(\asptv)$ $=$ $\semv(\agroundp{\asppv})$ by definition, while $\semv(\agroundp{\asppv})$ $=$ $\semv(\grounddlp{\asppv})$ follows from the fact that the AFs induced by $\agroundp{\asppv}$ and $\grounddlp{\asppv}$ are the same according to Lemma~\ref{lem:grd1}.
\end{proof}

\subsection{Proofs for Section~\ref{subsec:simp1}}

For the proofs for the Theorems of Section~\ref{subsec:simp1} we first introduce additional notation.  To start, for an argumentation theory $\asptv$ we use $\app$ to denote the approximated predicates of $\asptv$ (see Definition~\ref{def:approx}), while $\nap$ denotes the non-approximated predicates, and $\prp = \app \cup \nap$.

Moreover, let $\datpva$ be the Datalog program associated to the theory $\asptv$ as defined by Transformation~\ref{def:transf2}.  Then, $\appx$ denotes the auxiliary predicates in $\datpva$ used for rules of $\asptv$ defining approximated predicates, \ie~$\appx = \{\auxnr \mid \hedv \in \app, \hedv \textit{ occurs in the head of a } \rulv \in \rul \} \cup \{\namv \mid \hedv \in \app, \hedv \textit{ occurs in the head of a } \rulv \in \ruld, \pfua{\rulv} = \namv  \}$. Analogously, $\napx$ denotes the auxiliary predicates used in $\datpva$ for rules of $\asptv$ defining non-approximated predicates. With these notations we can state lemmas~\ref{lem:auxgrd2.2.1}-\ref{lem:auxgrd2.2.3}, which play the same role for Theorem~\ref{thm:grd2}, that Lemma~\ref{lem:auxgrd2} plays for Theorem~\ref{thm:grd1} in Section~\ref{ap:prfsgrd1}.

\begin{lemma}\label{lem:auxgrd2.2.1} Let $\asppv = \aspp$ be an \argth~and $\datpva$ de Datalog program for $\asppv$ as per Transformation~\ref{def:transf2}.   Then, {\bf i)} $\argv \in \artsapr{\agroundp{\asppv}}{\nap}$ iff $\derg{\argv}{1} \in \derspr{\groundp{\datpva}}{\napx}$; {\bf ii)}   $\argv \in \artsapr{\agroundp{\asppv}}{\nap}$ with $\conc{\argv} = \stmv$ iff $\derg{\argv}{2} \in \derspr{\groundp{\datpva}}{\nap}$ with $\concd{\derg{\argv}{2}} = \stmv$; {\bf iii)} $\argv \in \artsapr{\agroundp{\asppv}}{\nap}$ with $\pfua{\trl{\argv}} = \namv$ iff $\derg{\argv}{3} \in \derspr{\groundp{\datpva}}{\napx}$ with $\concd{\derg{\argv}{3}} = \namv$.    
\end{lemma}

\begin{proof} We prove ii); the proofs for i) and iii) are similar. 

    Let $\leq$ be a total order on the predicates of the Datalog program $\datpva$ for $\asptv$ witnessing it to be stratifiable.  Let $\dpvt_1,\ldots,\dpvt_n$ be the corresponding stratification of $\datpva$. With some abuse of notation we use $\dpvt_i$ to denote also the predicates defined in $\dpvt_i$. 
    
    We do the proof by strong induction on $\dpvt_i$. Specifically, we prove for each $i \geq 0$: $\argv \in \artsapr{\agroundp{\asppv}}{\nap \cap \dpvt_i}$ iff $\derg{\argv}{2} \in \derspr{\groundp{\datpva}}{\nap \cap \dpvt_i}$ ($\concd{\derg{\argv}{2}} = \stmv$ is clear from the structure of $\derg{\argv}{2}$).
    
    (Base case) For the proof of the claim for $i=0$, note first that the program $\dpvt_0$ must be free of negation by the definition of stratification.  But by the definition of $\datpva$ (in particular, Equation~\ref{eq:transf2_1} and the definition of the transformation for assumptions) this implies every predicate in $\dpvt_0 \cap \prp$ depends negatively only on approximated predicates.  But then, since non-approximated predicates can only depend on non-approximated predicates by Definition~\ref{def:approx}, every predicate in $\dpvt_0 \cap \nap$ depends only positively on other (non-approximated) predicates.  The latter implies that arguments in $\artpr{\agroundp{\asppv}}{\nap \cap \dpvt_0}$ are those which are not attacked by any argument, and, thus, $\artsapr{\agroundp{\asppv}}{\nap \cap \dpvt_0}$ $=$ $\artpr{\agroundp{\asppv}}{\nap \cap \dpvt_0}$.  Now, because $\dpvt_0$ is free of negation, that $\argv \in \artsapr{\agroundp{\asppv}}{\nap \cap \dpvt_0}$ $=$ $\artpr{\agroundp{\asppv}}{\nap \cap \dpvt_0}$ iff $\derg{\argv}{2} \in \derspr{\groundp{\datpva}}{\nap \cap \dpvt_0}$ then follows, as in the case of Lemma~\ref{lem:auxgrd2}, from the structure of the arguments and derivations together with Lemma~\ref{lem:auxgrd1}.  
    
    (Inductive case, $\Rightarrow$) Let us now prove the claim for $i \geq 1$ assuming we have proven it for each $j$ \stt~$i > j \geq 0$. So let $\argv \in \artsapr{\agroundp{\asppv}}{\nap \cap \dpvt_i}$ and consider the derivation $\derg{\argv}{2}$. Clearly, the only way $\derg{\argv}{2} \not\in \derspr{\groundp{\datpva}}{\nap \cap \dpvt_i}$ is that there is a $\hrulvp{1}\sigma \in \derg{\argv}{2}$ or $\hkvarp{1} \in \derg{\argv}{2}$ ($\kvar \in \kbso$)  \stt~some $l_k(\vec{X_k})\sigma$ ($1 \leq k \leq n$, $\l_k \in \nap$) (as in Equation~\ref{eq:transf2_1} and the definition of the transformation for assumptions respectively) can be derived from $\datpva$.  For simplicity let us assume $\hrulvp{1}\sigma \in \derg{\argv}{2}$; the case $\hkvarp{1} \in \derg{\argv}{2}$ is identical.  Because of the structure of $\datpva$ and stratification this means that there is a $\derg{\argv'}{2} \in \derspr{\groundp{\datpva}}{\nap \cap \dpvt_j}$ for $j < i$ with $\concd{\derg{\argv'}{2}} = l_k(\vec{X_k})\sigma$. From this, by inductive hypothesis we get that $\argv' \in \artsapr{\agroundp{\asppv}}{\nap \cap \dpvt_j}$.  But then (because $\conc{\argv'} =l_k(\vec{X_k})\sigma \in \cona{\wpa{\rulv\sigma}}$ for $\rulv\sigma \in \rulsa{\argv}$) $\argv'$ attacks $\argv$, which contradicts the assumption that $\argv \in \artsapr{\agroundp{\asppv}}{\dpvt_i}$.  Thus, $\derg{\argv}{2} \in \derspr{\groundp{\datpva}}{\nap \cap \dpvt_i}$.  
    
    (Inductive case, $\Leftarrow$) Let now, on the other hand, $\derg{\argv}{2} \in \derspr{\groundp{\datpva}}{\nap \cap \dpvt_i}$ and consider the argument $\argv \in \artpr{\agroundp{\asppv}}{\nap \cap \dpvt_i}$.  Consider now also any attacker $\argv' \in \arts{\agroundp{\asppv}}$ of $\argv$. Then  $\conc{\argv'} = l_k(\vec{X_k})\sigma \in \cona{\wpa{\rulv\sigma}}$ for $\rulv\sigma \in \rulsa{\argv}$ or $\conc{\argv'} = l_k(\vec{t_k}) \in \cona{\kvar}$ for $\kvar \in \pre{\argv} \cap \kbso$.  For simplicity, let us again assume $l_k(\vec{X_k})\sigma \in \cona{\wpa{\rulv\sigma}}$; the case $l_k(\vec{t_k}) \in \cona{\kvar}$ is identical.   Since the predicate in $\concd{\derg{\argv}{2}} = \conc{\argv} \in \nap$, also $l_k \in \nap$.  Then $l_k(\vec{X_k})\sigma$ occurs negatively in $\hrulvp{1}\sigma \in \derg{\argv}{2}$, which means, by stratification, that $l_k \in \nap \cap \dpvt_j$ for $j < i$.  But since $\derg{\argv}{2} \in \derspr{\groundp{\datpva}}{\nap \cap \dpvt_i}$, there must be a $\derg{\argv''}{2} \in \derspr{\groundp{\datpva}}{\nap \cap \dpvt_q}$ with $q < j$ and $\concd{\derg{\argv''}{2}} = l_{k'}'(\vec{X_{k'}})\sigma'$ occurring negatively in a $\hatp{\rulv'}{1}\sigma' \in \derg{\argv'}{2}$.  By inductive hypothesis, then $\argv'' \in \artsapr{\agroundp{\asppv}}{\nap \cap \dpvt_q}$. In other words, every attacker $\argv'$ of $\argv$ is, in turn, attacked by some $\argv'' \in \artsapr{\agroundp{\asppv}}{\nap \cap \dpvt_q}$ for a $q < i$ and, thus, $\argv \in \artsapr{\agroundp{\asppv}}{\nap \cap \dpvt_i}$. This concludes the inductive case and, thus, the proof. 
\end{proof}

\begin{lemma}\label{lem:auxgrd2.2.2} Let $\asppv = \aspp$ be an \argth~and $\datpva$ de Datalog program for $\asppv$ as per Transformation~\ref{def:transf2}.   Then, {\bf i)} if  $\argv \in \artse{\agroundp{\asppv}}$, then $\derg{\argv}{1} \in \ders{\groundp{\datpva}}$; {\bf ii)}   if $\argv \in \artse{\agroundp{\asppv}}$ with $\conc{\argv} = \stmv$, then $\derg{\argv}{2} \in \ders{\groundp{\datpva}}$ with $\concd{\derg{\argv}{2}} = \stmv$; {\bf iii)} if  $\argv \in \artse{\agroundp{\asppv}}$ with $\pfua{\trl{\argv}} = \namv$, then $\derg{\argv}{3} \in \ders{\groundp{\datpva}}$ with $\concd{\derg{\argv}{3}} = \namv$.     
\end{lemma}

\begin{proof} We prove ii); the proofs of i) and iii) are similar. Let $\argv \in \artse{\agroundp{\asppv}}$ and consider the derivation $\derg{\argv}{2}$. Clearly, the only way $\derg{\argv}{2} \not\in \ders{\groundp{\datpva}}$ is that there is a $\hrulvp{1}\sigma \in \derg{\argv}{2}$ or $\hkvarp{1} \in \derg{\argv}{2}$ ($\kvar \in \kbso$)  \stt~some $l_k(\vec{X_k})\sigma$ ($1 \leq k \leq n$, $\l_k \in \nap$) (as in Equation~\ref{eq:transf2_1} and the definition of the transformation for assumptions respectively) can be derived from $\datpva$.  For simplicity let us assume $\hrulvp{1}\sigma \in \derg{\argv}{2}$; the case $\hkvarp{1} \in \derg{\argv}{2}$ is identical.  Because of the structure of $\datpva$ and stratification this means that there is a $\derg{\argv'}{2} \in \derspr{\groundp{\datpva}}{\nap}$ with $\concd{\derg{\argv'}{2}} = l_k(\vec{X_k})\sigma$. From this, by Lemma~\ref{lem:auxgrd2.2.1} (item ii), we get that $\argv' \in \artsa{\agroundp{\asppv}}$.  But then (because $\conc{\argv'} =l_k(\vec{X_k})\sigma \in \cona{\wpa{\rulv\sigma}}$ for $\rulv\sigma \in \rulsa{\argv}$) $\argv'$ attacks $\argv$, which contradicts the assumption that $\argv \in \artse{\agroundp{\asppv}}$.  Thus, $\derg{\argv}{2} \in \ders{\groundp{\datpva}}$.  That also $\concd{\derg{\argv}{2}} = \stmv$ is clear from the structure of $\derg{\argv}{2}$.   \end{proof}

\begin{lemma}\label{lem:auxgrd2.2.3} Let $\asppv = \aspp$ be an \argth~and $\datpva$ de Datalog program for $\asppv$ as per Transformation~\ref{def:transf2}.   Then, {\bf i)} if $\derg{\argv}{1} \in \ders{\groundp{\datpva}}$, then $\argv \in \arts{\agroundp{\asppv}}$; {\bf ii)}   if $\derg{\argv}{2} \in \ders{\groundp{\datpva}}$ with $\concd{\derg{\argv}{2}} = \stmv$, then $\argv \in \arts{\agroundp{\asppv}}$ with $\conc{\argv} = \stmv$; {\bf iii)} if $\derg{\argv}{3} \in \ders{\groundp{\datpva}}$ with $\concd{\derg{\argv}{3}} = \namv$, then $\argv \in \arts{\agroundp{\asppv}}$ with $\pfua{\trl{\argv}} = \namv$.     
\end{lemma}

\begin{proof} This follows, as in the case of Lemma~\ref{lem:auxgrd2}, from Lemma~\ref{lem:auxgrd1} as well as the structure of the arguments and derivations in the statement of the lemma. \end{proof}

The following lemma is now the analogue of Lemma~\ref{lem:auxgrd3} in Section~\ref{ap:prfsgrd1} for Transformation~\ref{def:transf2}. 

\begin{lemma} \label{lem:auxgrd3.2} Let $\asppv = \aspp$ be an \argth, $\rulv \in \rul$, $\sigma: \vars \mapsto \heru{\asppv}$, $s \in \preds$ occurring in $\asppv$, and $\datpva$ the Datalog program for $\asppv$ as per Transformation~\ref{def:transf2}.   
Then,  {\bf i)} $\auxt{\rulv}{\xlv}\sigma \in \gqr{\datpva}{\auxnr}$ iff there is a $\derg{\argv}{1} \in \ders{\groundp{\datpva}}$ for an 
$\argv \in \arts{\agroundp{\asppv}}$ with $\trl{\argv} = \rulv\sigma$; {\bf ii)} $s({\xlv})\sigma \in \gqr{\datpva}{s}$ iff there is a 
$\derg{\argv}{2} \in \ders{\groundp{\datpva}}$ or $\derg{\argv}{3} \in \ders{\groundp{\datpva}}$  for an $\argv \in \arts{\agroundp{\asppv}}$ 
with $\conc{\argv} = s({\xlv})\sigma$ or $\pfua{\trl{\argv}} = s({\xlv})\sigma$.  
\end{lemma}

\begin{proof} This follows from the definitions of the queries ($\gqr{\datpva}{\auxnr}$ and $\gqr{\datpva}{s}$), the construction of $\datpva$, and Lemma~\ref{lem:auxgrd2.2.3}.\end{proof}

\begingroup
\renewcommand{\thelemma}{2} 
\begin{lemma}
Let $\asppv$ be an \argth~and $\grounddlp{\asppv}$ the grounding via the Datalog program $\datpva$ as per Transformation~\ref{def:transf2}. Then, $\artse{\agroundp{\asppv}} \subseteq \arts{\grounddlp{\asppv}} \subseteq \arts{\agroundp{\asppv}}$ and $\atts{\grounddlp{\asppv}} = \aproj{\atts{\agroundp{\asppv}}}{\arts{\grounddlp{\asppv}}} $.
\end{lemma}
\endgroup    

\begin{proof} 
Let $\asppv = \aspp$.    

($\artse{\agroundp{\asppv}} \subseteq \arts{\grounddlp{\asppv}} \subseteq \arts{\agroundp{\asppv}}$)  We have that $\rulv\sigma \in \grounddlp{\rul}$ for a $\rulv \in \rul$ iff $\auxt{\rulv}{\xlv}\sigma \in \gqr{\datpva}{\auxnr}$ iff (by Lemma~\ref{lem:auxgrd3.2} (i)) there is a $\derg{\argv}{1} \in \ders{\groundp{\datpva}}$ for an $\argv \in \arts{\agroundp{\asppv}}$ with $\trl{\argv} = \rulv\sigma$.  Moreover, by Lemma~\ref{lem:auxgrd2.2.2} (i) and Lemma~\ref{lem:auxgrd3.2} (i), each $\rulv\sigma$ for an $\argv \in \artse{\agroundp{\asppv}}$ with $\trl{\argv} = \rulv\sigma$ is in $\grounddlp{\rul}$. Since each $\rulv\sigma \in \rulsa{\artse{\agroundp{\asppv}}}$ is the top-rule of some $\argv \in \artse{\agroundp{\asppv}}$ (the set of arguments is closed under sub-arguments), this means every $\rulv\sigma \in \rulsa{\artse{\agroundp{\asppv}}}$ is also in $\grounddlp{\rul}$.  By similar reasoning but making use of lemmas~\ref{lem:auxgrd3.2} (ii) and~\ref{lem:auxgrd2.2.2} (ii), we have that if $b \in \grounddlp{\kbso}$ then $b \in \arts{\agroundp{\asppv}}$, while if $b \in \artse{\agroundp{\asppv}} \cap \kbso$ then also $b \in \grounddlp{\kbso}$.   But then, we have concluded that $\grounddlp{\rul} \subseteq \rulsa{\arts{\agroundp{\asppv}}}$ and $\grounddlp{\kbso} \subseteq \arts{\agroundp{\asppv}}$, while $\grounddlp{\kbsa} = \arts{\agroundp{\asppv}} \cap \kbsa$ by definition.  Thus, every argument that can be constructed in $\grounddlp{\asppv}$ can also be constructed in $\agroundp{\asppv}$.  In the same manner, we have seen that $\rulsa{\artse{\agroundp{\asppv}}} \subseteq \grounddlp{\rul}$ and $(\artse{\agroundp{\asppv}} \cap \kbso) \subseteq \grounddlp{\kbsa}$, while  $\grounddlp{\kbsa} = \arts{\agroundp{\asppv}} \cap \kbsa = \artse{\agroundp{\asppv}} \cap \kbsa$ by definition.  Thus, every argument in  $\artse{\agroundp{\asppv}}$ can also be constructed in $\grounddlp{\asppv}$.

($\aproj{\atts{\agroundp{\asppv}}}{\arts{\grounddlp{\asppv}}} \subseteq \atts{\grounddlp{\asppv}}$) Now consider an $\argv \in \arts{\grounddlp{\asppv}}$ that attacks an $\argv' \in \arts{\grounddlp{\asppv}}$ in $\agroundp{\asppv}$. This means $\conc{\argv} \in S({\xlv})\sigma$ for a $s({\xlv})\sigma \in \wpa{\argv'}$ and $c: \acrul{\astmv{\xlv}}{\astmsv{\xlv}} \in \con$. But this means also there is a $\argv'' \in \subo{\argv'}$ with $\conc{\argv''} = s({\xlv})\sigma$ or $\pfua{\trl{\argv''}} = s({\xlv})\sigma$.  Moreover, since $\argv' \in \arts{\grounddlp{\asppv}}$, by Lemma~\ref{lem:auxgrd3.2}, $\derg{\argv'}{1} \in \ders{\groundp{\datpva}}$.  From this it is clear (since $\derg{\argv''}{1} \subseteq \derg{\argv'}{1}$) that then also $\derg{\argv''}{2} \in \ders{\groundp{\datpva}}$ or $\derg{\argv''}{3} \in \ders{\groundp{\datpva}}$ from where it follows, again by  Lemma~\ref{lem:auxgrd3.2}, that $s({\xlv})\sigma \in \gqr{\datpva}{s}$ and, hence, $c\sigma \in \grounddlp{\con}$.  We thus get that the attack from $\argv$ to $\argv'$ is also in $\atts{\grounddlp{\asppv}}$. 

($\atts{\grounddlp{\asppv}} \subseteq \aproj{\atts{\agroundp{\asppv}}}{\arts{\grounddlp{\asppv}}}$) That attacks between arguments $\argv,\argv'$ in $\grounddlp{\asppv}$ are preserved in $\agroundp{\asppv}$ follows, as in the proof of Lemma~\ref{lem:grd1}, from the fact that such attacks are obtained via a $c\sigma \in \grounddlp{\con}$ for $c \in \con$ and $\sigma: \vars \mapsto \heru{\datpva}$. But, since $\heru{\datpva} = \heru{\asppv}$, then also $c\sigma \in \agroundp{\con}$ and, thus, $\argv$ also attacks $\argv'$ in $\agroundp{\asppv}$.\end{proof}

\begingroup
\renewcommand{\thetheorem}{2} 
\begin{theorem}
Let $\asppv$ be an \argth~and $\grounddlp{\asppv}$ the grounding via the Datalog program $\datpva$ as per Transformation~\ref{def:transf2}. Then, $\semv(\asptv)$ $=$ $\semv(\agroundp{\asppv})$ $=$ $\semv(\grounddlp{\asppv})$ for $\semv \in \semlistc$.
\end{theorem}
\endgroup

\begin{proof}

Note that by Lemma~\ref{lem:grd2} the AF induced by $\grounddlp{\asppv}$ is obtained by taking the AF induced by $\agroundp{\asppv}$ and removing a (potentially empty) set of arguments $\argsv \subseteq \arts{\agroundp{\asppv}} \setminus \artse{\agroundp{\asppv}}$ as well as attacks involving arguments in $\argsv$.  

We first prove $\com(\agroundp{\asppv})$ $=$ $\com(\grounddlp{\asppv})$.

($\com(\agroundp{\asppv})$ $\subseteq$ $\com(\grounddlp{\asppv})$) Taking a complete extension $E$ of $\agroundp{\asppv}$, first of all we have that also $E \subseteq \arts{\grounddlp{\asppv}}$ because $E \subseteq \artse{\agroundp{\asppv}}$ by the fact that $E$ is a complete extension. Secondly, $E$ is also conflict-free in $\grounddlp{\asppv}$ as $E \subseteq \arts{\grounddlp{\asppv}}$ and the attacks between arguments in E in $\grounddlp{\asppv}$ are exactly as in $\agroundp{\asppv}$.  As for admissibility of $E$ in $\grounddlp{\asppv}$, there are no new attacks in $\grounddlp{\asppv}$ (w.r.t those in $\agroundp{\asppv}$) so no new arguments that need to be defended by $E$.  On the other hand, defenders in $E$ are also preserved in $\grounddlp{\asppv}$ as the only arguments of $\agroundp{\asppv}$ that are not arguments of $\grounddlp{\asppv}$ are arguments that are not tentative and these do not form part of any extension of $\agroundp{\asppv}$ (and, hence, also not of $E$). 

($\com(\grounddlp{\asppv})$ $\subseteq$ $\com(\agroundp{\asppv})$) Take now, on the other hand, a complete extension $E$ of $\grounddlp{\asppv}$.  $E$ is also conflict-free in $\agroundp{\asppv}$ as the only attacks in which $\agroundp{\asppv}$ differs from $\grounddlp{\asppv}$ is in attacks involving $\argsv$, but arguments of $\argsv$ will not be part of $E$ by the fact that they are not arguments of $\grounddlp{\asppv}$.  As to admissibility, again we need to consider only attacks to $E$ in $\agroundp{\asppv}$ by arguments in $\argsv$.  But each such argument in $\argsv$ is attacked by some certain argument $\argv$ of $\agroundp{\asppv}$ by definition.  But clearly also the set of all certain arguments $C$ of $\agroundp{\asppv}$ is also in $E$.  Indeed, take any set of arguments $\argsv'$ attacking $C$ in $\grounddlp{\asppv}$. Then, because these arguments $\argsv'$ are also arguments of $\agroundp{\asppv}$, $C$ can defend itself from $\argsv'$ without $\argsv'$ being able to defend itself from $C$ (i.e. there is an argument in $\argsv'$ which is not defended by $\argsv'$ from $C$).  This holds also for the set $E$ and, thus, $E$ does not attack $C$ but also $C$ does not attack $E$ (as in that latter case $E$ would not be admissible).  Thus, actually $C \subseteq E$ because $E$ is complete.  In conclusion, we have that $E$ is also admissible in $\agroundp{\asppv}$.  
 
Finally, $E$ is also complete in $\agroundp{\asppv}$, because if there were a $\argv \in \artse{\aground{\asppv}} \setminus E$ s.t. $E \cup \{\argv\}$ is complete, then the same would hold for $\grounddlp{\asppv}$, which contradicts the completeness of $E$ in $\grounddlp{\asppv}$. 

Now, since we have seen that $\com(\agroundp{\asppv})$ $=$ $\com(\grounddlp{\asppv})$, clearly also the grounded and preferred extensions coincide (these select the unique minimal and the maximal complete extensions respectively).  As for the stable semantics, we have seen in the proof for the complete semantics, that certain arguments are part of all complete extensions in both $\agroundp{\asppv}$ and $\grounddlp{\asppv}$.  But this means also that the (non-tentative) arguments $\argsv$ in which $\agroundp{\asppv}$ and $\grounddlp{\asppv}$ differ are attacked by every complete extension in both $\agroundp{\asppv}$ and $\grounddlp{\asppv}$.  Thus, also the stable extensions coincide.\end{proof}

We finally finish up the proofs by giving those for Lemma~\ref{lem:grd3} and Theorem~\ref{thm:grd3}.

\begingroup
\renewcommand{\thelemma}{3} 
\begin{lemma}
Let $\asppv$ be an \argth, $\grounddlph{\asppv}$ the grounding of $\asppv$ via Algorithm~\ref{alg:ground_theory}, and $\grounddlpt{\asppv}$ the grounding via the Datalog program $\datpva$ as per Transformation~\ref{def:transf2}\footnote{I.e. without the further simplifications in Algorithm~\ref{alg:ground_theory}.}. Then, $\{\conc{E} \mid E \in \semv(\grounddlpt{\asppv})\}$ $=$ $\{\conc{E} \mid E \in \semv(\grounddlph{\asppv})\}$ for $\semv \in \semlistc$.
\end{lemma}
\endgroup

\begin{proof} We start by proving $\{\conc{E} \mid E \in \com(\grounddlpt{\asppv})\}$ $=$ $\{\conc{E} \mid E \in \com(\grounddlph{\asppv})\}$.  Note that the only difference between $\grounddlpt{\asppv}$ and $\grounddlph{\asppv}$ (as can be deduced, in particular, from lines 1-18 of Algorithm~\ref{alg:ground_theory}) is that $\grounddlph{\ruls} \subseteq \grounddlpt{\ruls}$, while $\grounddlpt{\kbsa} \subseteq \grounddlph{\kbsa}$.  More specifically, $\rulv \in \grounddlpt{\ruls}\setminus \grounddlph{\ruls}$ iff $\hedvr{\rulv} \in \grounddlph{\kbsa} \setminus \grounddlpt{\kbsa}$. Moreover, a $\hedvr{\rulv} \in \grounddlph{\kbsa} \setminus \grounddlpt{\kbsa}$ whenever $\hedvr{\rulv}$ can be derived solely using facts and strict rules in $\grounddlpt{\asppv}$.  I.e. there is a $\argv_{\hedvr{\rulv}} \in \arts{\grounddlpt{\asppv}}$ with $\conc{\argv_{\hedvr{\rulv}}} = \hedvr{\rulv}$ and $\pre{\argv_{\hedvr{\rulv}}} \subseteq \grounddlpt{\kbsa}$, while $\rulsa{\argv_{\hedvr{\rulv}}} \subseteq \grounddlpt{\ruls}$.  But this means that for any argument $\argv$ of $\grounddlpt{\asppv}$ where a $\rulv \in \grounddlpt{\ruls}\setminus \grounddlph{\ruls}$ is used, the subargument $\argv'$ of $\argv$ with $\trl{\argv'} = \rulv$ can be replaced in $\grounddlph{\asppv}$ with the argument $\argv'' = \hedvr{\rulv} \in \grounddlph{\kbsa}$. On the other hand, for any argument $\argv$ of $\grounddlph{\asppv}$ where a $\hedvr{\rulv} \in \grounddlph{\kbsa} \setminus \grounddlpt{\kbsa}$ is used, can be replaced in  $\grounddlpt{\asppv}$ with a $\argv_{\hedvr{\rulv}} \in \arts{\grounddlpt{\asppv}}$ as described above.  
    
Moreover, attacks are preserved when converting arguments of  $\grounddlpt{\asppv}$ to arguments of $\grounddlph{\asppv}$ and viceversa in the described manner.  The reason for this is that, first of all, for any argument in any of the groundings with conclusion $c$, there is a corresponding argument with the same conclusion in the other grounding.  Secondly, the translations only involve substituting "strict" arguments, i.e. those making use only of facts and strict rules, with other strict arguments.  Thus, the conversion only affects (sub-) arguments that are not attacked.   Thirdly, $\grounddlph{\con} = \grounddlpt{\con}$ by Algorithm~\ref{alg:ground_theory} (line 21).

To conclude the proof for the complete semantics, note that any  $E \in \com(\grounddlpt{\asppv})$ can be converted to an $E' \in \com(\grounddlph{\asppv})$ and viceversa again by first replacing (sub-) arguments as described above.  Since the translation only substitutes strict arguments for strict arguments, the (sub-) arguments introduced in the transformation are not attacked by any argument and thus admissibility is preserved by the translation. Moreover, the transformation as described above can be used for any argument and will thus produce, for each argument in one extension, a corresponding argument in the other extension with the same conclusion.   Finally, completeness is preserved as can be seen by the following argument. Consider first that $E$ obtained by transforming $E'$ is not complete.  This means that there is an $\argv \in \arts{\grounddlpt{\asppv}}$ s.t. $\argv \not\in E$ and $E \cup \{\argv\}$ is also admissible. Then we can transform $\argv$ into an $\argv' \in \arts{\grounddlph{\asppv}}$ as above and, since $\argv'$ is different to $\argv$ only in the strict parts (and these do not affect attacks), we have that $E' \cup \{\argv'\}$ is also admissible, which contradicts that $E' \in \com(\grounddlph{\asppv})$.  Thus, $E \in \com(\grounddlph{\asppv})$.  The argument for $E' \in \com(\grounddlph{\asppv})$ from $E \in \com(\grounddlpt{\asppv})$ is analogous.

Now we consider the grounded semantics. We have seen that for a grounded extension $E$ of $\grounddlph{\asppv}$ there is (because grounded extensions are also complete) a complete extension $E'$ of $\grounddlpt{\asppv}$ with same conclusions.  Now if $E'$ is not grounded in $\grounddlpt{\asppv}$, then consider the grounded extension $E''$ of $\grounddlpt{\asppv}$.  If $E''$ has the same conclusions as $E'$ then we are done.  Otherwise, $E''$ must have fewer conclusions than $E'$ but then there is also a complete extension $E'''$  in $\grounddlph{\asppv}$ with the same conclusions as $E''$ and, thus, fewer conclusions than $E$.  But this means that there is some argument in $E$ that is not in $E''$ and, thus, $E$ is not minimal complete and, thus, also not grounded.  This is a contradiction and, therefore, $E'$ is grounded.  The arguments for the other direction and, then, the preferred semantics are analogous. 

As for the stable semantics, consider a stable extension $E$ of $\grounddlph{\asppv}$.  Then, again (by the fact that stable extensions are complete), there is a complete extension $E'$ of $\grounddlpt{\asppv}$ with same conclusions.  Now, assume $E'$ is not stable; i.e. there is an argument $\argv$ of $\grounddlpt{\asppv}$ that is not in $E'$ and that $E'$ does not attack.   But then we can transform the argument $\argv$ into an argument $\argv'$ of $\grounddlph{\asppv}$ using the transformation we have described above.  Since this transformation only involves substituting strict (sub-) arguments (i.e. those involving only facts and strict rules and, hence, that are not attacked by any argument) for strict arguments and the conclusions of $E$ and $E'$ are the same (i.e. outgoing attacks are also preserved), then $\argv'$ is also not attacked by $E$ which is a contradiction with the stability of $E$.  Thus, we have that $E'$ must also be stable.  The proof for the other direction is, again, analogous.\end{proof}

\begingroup
\renewcommand{\thetheorem}{3} 
\begin{theorem}\sloppy Let $\asppv$ be an \argth, $\grounddlph{\asppv}$ the grounding of $\asppv$ via Algorithm~\ref{alg:ground_theory}. Then, $\{\conc{E} \mid E \in \semv(\asptv)\}$ $=$ $\{\conc{E} \mid E \in \semv(\agroundp{\asppv})\}$ $=$ $\{\conc{E} \mid E \in \semv(\grounddlp{\asppv})\}$ for $\semv \in \semlistc$. 
\end{theorem}

\begin{proof} Follows from Lemma~\ref{lem:grd3} and Theorem~\ref{thm:grd2}. \end{proof}

\endgroup

\end{document}